\documentclass[10pt,twocolumn,letterpaper]{article}

\usepackage{iccv}
\usepackage{times}
\usepackage{epsfig}
\usepackage{graphicx}
\usepackage{amsmath}
\usepackage{amssymb}
\usepackage{algorithm}
\usepackage{algorithmicx}
\usepackage{algpseudocode}

\usepackage{mathtools}
\usepackage{caption}
\usepackage{subcaption}
\usepackage{float}
\usepackage{placeins}

\usepackage{amsthm}

\newcommand*\diff{\mathop{}\!\mathrm{d}}

\DeclareMathOperator*{\argmin}{arg\,min}
\newcommand{\mbeq}{\overset{!}{=}}
\newcommand{\Lagr}{\mathcal{L}}
\newtheorem{definition}{Definition}
\newtheorem{theorem}{Theorem}
\newtheorem{proposition}{Proposition}
\long\def\/*#1*/{}

\usepackage[breaklinks=true,bookmarks=false]{hyperref}

\iccvfinalcopy 

\def\iccvPaperID{6578} 

\begin{document}

\title{Analyzing the Variety Loss in the Context of Probabilistic Trajectory Prediction}

\author{Luca Anthony Thiede \thanks {Work done during internship at Volkswagen Group of America Innovation and Engineering Center located in Belmont, California.}\\
 Georg-August-Universit\"at G\"ottingen \\
{\tt\small luca.thiede@yahoo.com}
\and
Pratik Prabhanjan Brahma\\
Innovation and Engineering Center California\\
Volkswagen Group of America\\
{\tt\small pratik.brahma@vw.com}
}

\maketitle

\begin{abstract}
   Trajectory or behavior prediction of traffic agents is an important component of autonomous driving and robot planning in general. It can be framed as a probabilistic future sequence generation problem and recent literature has studied the applicability of generative models in this context. The variety or Minimum over N (MoN) loss, which tries to minimize the error between the ground truth and the closest of N output predictions, has been used in these recent learning models to improve the diversity of predictions. In this work, we present a proof to show that the MoN loss does not lead to the ground truth probability density function, but approximately to its square root instead. We validate this finding with extensive experiments on both simulated toy as well as real world datasets. We also propose multiple solutions to compensate for the dilation to show improvement of log likelihood of the ground truth samples in the corrected probability density function. 
\end{abstract}

\section{Introduction}
\begin{figure}
    \centering
    \includegraphics[width=.8\linewidth]{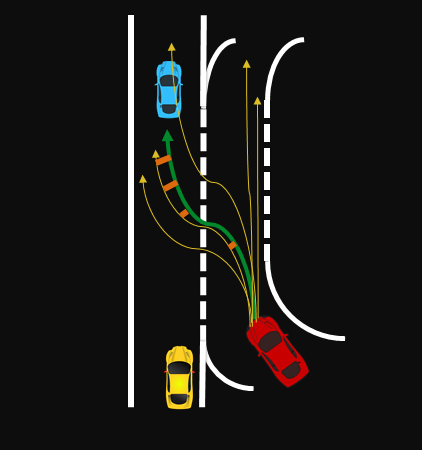}
    \caption{Trajectory prediction is a multimodal problem. In order to learn this distribution, the variety loss can be used. It is computed as the distance between the groundtruth (green) trajectory and its closest prediction.}
    \label{fig:highway}
\end{figure}

Trajectory prediction is an important problem with many applications. It can be used for tracking~\cite{application_tracking}, anomaly detection~\cite{application_anomaly}, video games~\cite{application_games} or safety simulation~\cite{application_simulation}. Arguably, the most safety critical application is to use trajectory prediction to help robots navigate environments that they share with other people, for example in the case of self driving cars. While driving, humans have an intuitive anticipation of what other traffic participants are likely to do and react accordingly. This is remarkable since future trajectories are non-deterministic and multimodal (See Figure \ref{fig:highway}).  
For this reason, a recent line of research takes the approach to model the natural probability distribution of recorded data, for example with mixture density networks~\cite{s_lstm, convolutional_tp}, occupancy grids~\cite{occupancy_grid} or generative models~\cite{sophie_gan, social_gan, vae_traj, traj_met_3}. One of the recent works, Social--GAN \cite{social_gan}, trained their generative model with a combination of the adversarial loss \cite{gan} and the variety loss (hereafter referred to as Minimum over N or MoN). They, and many other recently published works \cite{traj_met_1, traj_met_2, traj_met_3, traj_met_4, traj_met_5}, used the same MoN loss as a metric to benchmark their model against others, arguing that it is better suited to measure performance on multimodal data as compared to the widely used average displacement error \cite{ade_source}.
Additionally we noticed that researchers in other fields apart from trajectory prediction used variations of the MoN loss/metric as well \cite{original_paper, 3d_rec_met, traj_met_6, opt_flow_met, cui2018multimodal, bhattacharyya2018accurate, depth_map}. However, we could not find any theoretical analysis of the MoN loss/metric. \\
In this work, we present a proof in section \ref{sec:theory} that the optimal solution of the MoN loss is not the ground truth PDF but its square root instead. In section \ref{sec:metric}, we discuss if MoN can be used as a viable metric nonetheless. Then, in section \ref{sec:reover_gt}, we propose various algorithms to recover the true PDF from the learned one. We verify these results experimentally on simulated low--dimensional datasets in section \ref{sec:experiments_toy}. We also validate our hypotheses on a highway vehicles and a pedestrian trajectory prediction datasets by applying a compensating transformation on the distribution learned by MoN loss based generative model and show an improvement of ground truth samples in the sense of average marginalized log likelihood.\\
\section{Related work}
Trajectory prediction of traffic participants is a difficult problem. The model has to capture the many possible outcomes as well as the interactions of the person/vehicle to be predicted with other traffic participants and the environment. Early attempts of predicting the trajectories of humans under consideration of social interactions used a model of attractive and repulsive social forces~\cite{social_forces, social_forces2, scoial_forces3, social_forces_cool, ade_source} with promising results. Other approaches include using Gaussian processes \cite{trajectories_gaussian_processes} and continuum dynamics \cite{cont_crowd}.\\
Newer works are more data driven. Some \cite{s_lstm, convolutional_tp} use data to teach a network to predict the parameters of base distributions (Mixture Density Networks) \cite{mdn, convolutional_tp}. Others discretize the prediction space into a grid and predict the probability that one of these grid cells is occupied \cite{occupancy_grid, occupency2}. While these models show promising results, it is difficult to sample trajectories with a longer time horizon. This limitation is overcome by modeling longer trajectories directly using generative models. Generally these models learn to transform samples from a latent space into samples from a data distribution. The best known representatives for generative models are variational autoencoders (VAE) \cite{vae, vae_traj, traj_met_6} and generative adversarial networks (GANs) \cite{gan, social_gan, sophie_gan}. VAEs are trained by auto-encoding samples and optimizing a variational lower bound on the data distribution. GANs, on the other side, learn a discriminator jointly with the generator. The discriminator has the task to separate real data samples from generated ones while the generator has to produce samples that fool the discriminator. It was shown that this training procedure reaches the optimum if and only if the generator has learned the true data distribution \cite{gan}. Both models have seen successful applications on a wide array of tasks like texture synthesis \cite{texture_sythesis}, super resolution \cite{superresolution}, text to image synthesis \cite{text_to_image} or image synthesis from a mask \cite{image_synthesis_from_mask}.
MoN was originally introduced by \cite{original_paper} in the context of 3d point cloud generation and adopted by \cite{social_gan, traj_loss, cui2018multimodal} for trajectory prediction. Other works used MoN loss/metric or similar concepts for depth map prediction \cite{depth_map}, 3d reconstruction \cite{3d_rec_met}, activity prediction \cite{traj_met_6}, to improve the optimization of the variational lower bound in VAE \cite{bhattacharyya2018accurate} or for pixel flow prediction \cite{opt_flow_met}.

\section{The Minimum over N loss}
\label{sec:theory}
\begin{figure}
    \centering
    \includegraphics[width=1\linewidth]{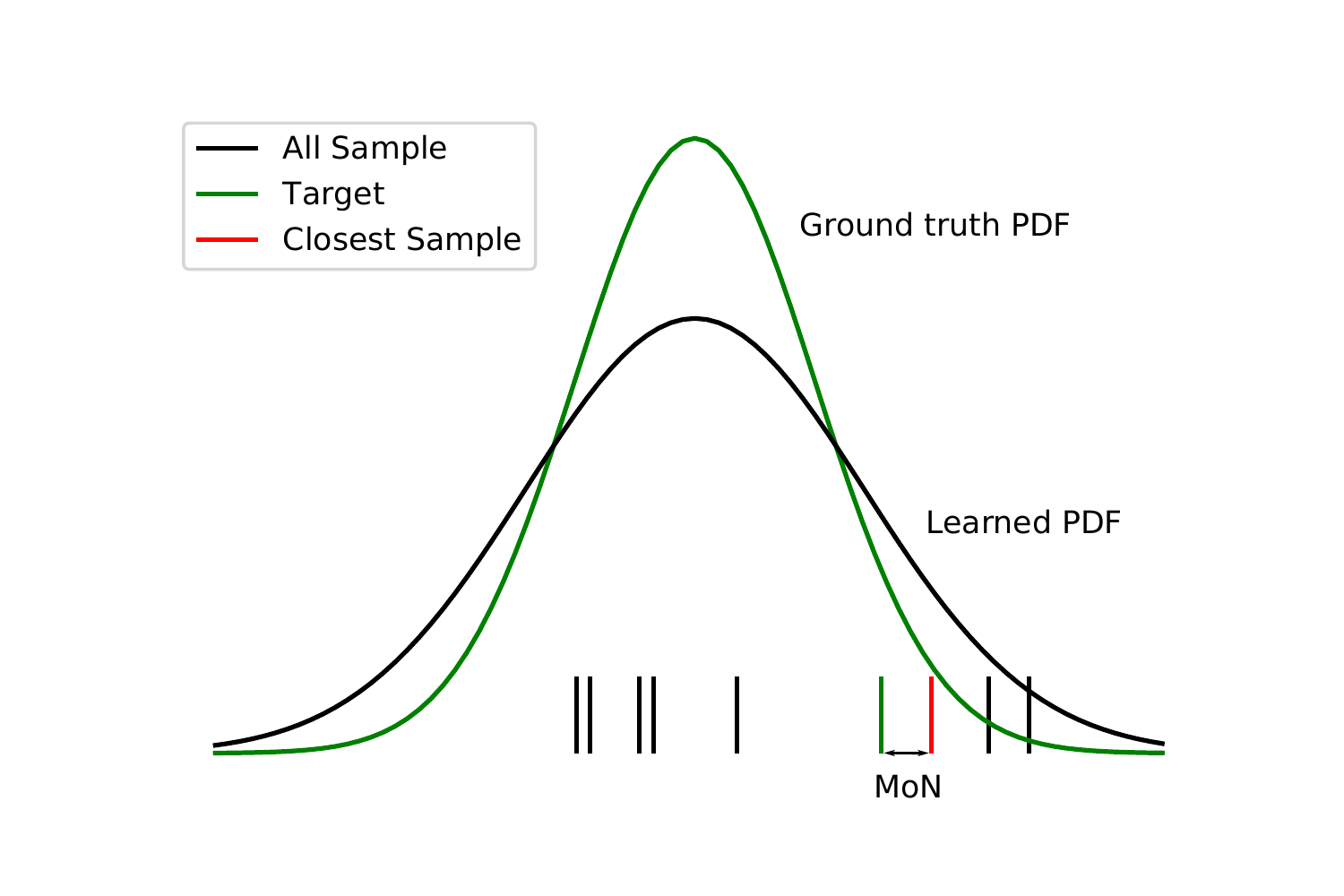}
    \caption{An illustration of the MoN loss in one dimension. Only the error of the sample with the smallest distance to the target sample is considered. This leads a model to learn the square root of the true PDF.}
    \label{fig:MoN_illustration}
\end{figure}

Given is a generative model $\boldsymbol{P(X|\text{I})}$, where $\boldsymbol{X} \in \mathbb{R}^n$ for some $n \in \mathbb{N}$ (for example $n=2T$ for 2 dimensional trajectories of length $T$) is the output to be generated and $\text{I}$ is a set of inputs. Then the MoN loss is defined as 
\begin{flalign}
    &\text{MoN}_P (\boldsymbol{x^*}) = \notag\\ 
    & \min_{\boldsymbol{x_1}, ..., \boldsymbol{x_N} \stackrel{\text{iid}}{\sim} P}\left(d(\boldsymbol{x^*},\boldsymbol{x_1}),d(\boldsymbol{x^*},\boldsymbol{x_2}),...,d(\boldsymbol{x^*},\boldsymbol{x_N})\right)
\end{flalign}
where $x^*$ is a ground truth sample and $\boldsymbol{x_1} \ldots \boldsymbol{x_N} \sim \boldsymbol{P(X|\text{I})}$ are samples generated from the model. The function $d(\cdot, \cdot)$ is some distance metric. One natural choice is the l2 distance $d(\boldsymbol{x}, \boldsymbol{y}) = ||\boldsymbol{x} - \boldsymbol{y}||_2$. An illustration in the one dimensional case is shown in Figure \ref{fig:MoN_illustration}.\\

In this paper we consider the question: given some ground truth probability distribution $P_T(x^*)$, does a model that was learned with the MoN loss converge towards this $P_T(x^*)$?
To get a better theoretical grasp, we consider the expectation value of the MoN loss:
\begin{definition}{EMoN:}
Given a probability density $P(\boldsymbol{x}): \mathbb{R}^n \rightarrow [0,1]$ and some point $\boldsymbol{x^*} \in \mathbb{R}^n$. Then we define the Expected-Minimum-over-N function 
\begin{align}
    &\text{EMoN}_P(\boldsymbol{x^*}) = \notag\\
    &\int \min\left(||\boldsymbol{x^*}-\boldsymbol{x_1}||_2,||\boldsymbol{x^*}-\boldsymbol{x_2}||_2,...,||\boldsymbol{x^*}-\boldsymbol{x_N}||_2\right) \notag\\ & \quad P(\boldsymbol{x_1})P(\boldsymbol{x_2})\ldots P(\boldsymbol{x_N}) \diff \boldsymbol{x_1} \diff \boldsymbol{x_2} \ldots \diff \boldsymbol{x_N} \label{eq:EMoN}
\end{align}
\end{definition}
We can estimate $\text{EMoN}_P(\boldsymbol{x^*})$ with 
\begin{align}
    & \widehat{\text{EMoN}}_P(\boldsymbol{x^*}) = \notag\\
    & \frac{1}{R} \sum^R \min_{\boldsymbol{x_1}, ..., \boldsymbol{x_N} \stackrel{\text{iid}}{\sim} P}(||\boldsymbol{x^*}-\boldsymbol{x_1}||_2,||\boldsymbol{x^*}-\boldsymbol{x_2}||_2,...,||\boldsymbol{x^*}-\boldsymbol{x_N}||_2) \label{eq:EEMoN}
\end{align}
where $R$ is the sample size for the expectation value. In the referenced literature $R$ in equation \eqref{eq:EEMoN} is set to $R = 1$. Since the variance of \ref{eq:EEMoN} is $\mathcal{O}(\frac{1}{R})$ one could question, if our theoretical results that are based on equation \ref{eq:EMoN} still hold. The experiments show though, that this is indeed the case. \\

Next, we can consider the expected MoN loss (MoN loss for short):
\begin{definition}{MoN loss:}
Given some target probability $P_T(x): \mathbb{R}^n \rightarrow [0,1]$ we define the Minimum-over-N loss as 
\begin{align}
    L_N(P_T, P) = \int P_T(\boldsymbol{x^*})  \text{EMoN}_P(\boldsymbol{x^*}) \diff x^* \label{eq:mon_loss}
\end{align}
\end{definition}
In a practical context, we would estimate this with samples from our dataset $\textbf{D}$:
\begin{align}
    \hat{L}_N(P_T, P) = \frac{1}{|\textbf{D}|}\sum_{x_{T}\in \textbf{D}}  \widehat{\text{EMoN}}_P(\boldsymbol{x^*})
\end{align}

The following theorem answers the question, whether a model trained with the MoN loss converges towards the true data distribution $P_T$:
\begin{theorem}
    For N big enough and $P_T$ differentiable with finite support, the differentiable PDF that minimizes the MoN loss is 
    \begin{align}
        \argmin_P{\hat{L}_N(P_T, P)} \approx \frac{\sqrt{P_T}}{C}
    \end{align}
    with some normalization constant $C$. \label{theorem} 
\end{theorem}
A proof is presented in the supplementary material. It is remarkable that this means that the MoN loss is a likelihood free loss (similar to the adversarial loss), that is, it does not assume any parametric form of the target distribution and can therefore be used to train a generative model.\\  

If $N=1$, there is a high chance that a model learns the mean of the known PDF. However with a large number of tries, it has the tendency to put more probability mass into regions with low ground truth probability. This is because putting more samples in areas with high probability will decrease the expected error only a little bit if there are already many samples, while, even if not likely, the prospect of a high error in the low probability area out weighs this decrease. This leads us to the following proposition:
\begin{proposition}
    Given $N_1 < N_2$, a ground truth PDF $P(\boldsymbol{x})$ and the family of PDFs
    \begin{align}
        P_k(\boldsymbol{x}) := \frac{1}{C_k} P(\boldsymbol{x})^k
    \end{align}
    Let $P_{k_i}(\boldsymbol{x})$ be the PDF out of this family, that minimizes $MoN_i$. Then 
    \begin{align}
        k_2 \le k_1
    \end{align}
    \label{proposition}
\end{proposition}
We verify this proposition in section \ref{sec:experiments_toy} experimentally. Proposition \ref{proposition} means, that only considering the family $P_k(x)$, the exponent $k(N)$ that minimizes MoN is monotonically falling with $N$. Because of Theorem \ref{theorem}, $k=0.5$ is a strict lower bound. Note, that this does not guarantee that a learner actually converges towards $P_k(\boldsymbol{x})$ (in fact it is easily seen that this is not the case for a multimodal distribution and $N=1$). We assume that the transformation that recovers the ground truth PDF from the PDF that minimizes MoN (we call this the compensation transformation from now on) belongs to the following family of transformations:
\begin{align}
    T_{\Bar{k}}(P(\boldsymbol{x})) = \frac{1}{ \int P(\boldsymbol{x})^{\Bar{k}} \diff x } P(\boldsymbol{x})^{\Bar{k}} \label{eq:transformation}
\end{align}
For some practical $N$ (where $\Bar{k} = \frac{1}{k}$), proposition \ref{proposition} gives us the intuition that $\Bar{k}$ is going to be less than $2$. \\

\section{MoN as a metric}
\label{sec:metric}
The ideal metric to compare probabilistic models would be a statistical divergence like the Kullback--Leibler divergence or the Jensen-–Shannon divergence between the learned and the ground truth distribution. Comparing the KL divergence is equivalent to comparing the log likelihood of the ground truth samples under the models. Since two dimensional trajectories with $T$ time steps in the future is $2T$ dimensional, estimating this log likelihood with generative models, where we do not have direct access to the likelihood of samples, is unfeasible for anything but very small $T$ (for small $T$ the learned PDF can be estimated by sampling from the model).\\

Recent work~\cite{social_gan} in trajectory prediction used MoN as a metric to compare their results against previous ones (\eg~\cite{s_lstm}). This can be problematic though (particularly if one of the models was trained with the MoN loss while the others were not), as it rewards a model that learned a less sharp distribution. Therefore, following~\cite{convolutional_tp} we advocate to use additionally to the MoN a second metric: The average log likelihood of the ground truth from the test set $\textbf{D}_\text{Test}$ under the marginalized learned distribution for every time step $t$:
\begin{align}
    &\Lagr^t_{\textbf{D}_\text{Test}}(P) = \notag\\  &\frac{1}{|\textbf{D}_\text{Test}|}\sum_{(x_{i,t}^*, y_{i,t}^*) \in \textbf{D}_\text{Test}} \log \int P(x_1, y_1, ..., x_{i,t}^*, y_{i,t}^*, ..., x_T, y_T) \notag\\ 
    & \diff x_1 \diff y_1,...,\diff x_{t-1}, \diff y_{t-1}, \diff x_{t+1}, \diff y_{t+1}, ..., \diff x_T, \diff y_T \label{eq:marginalized_distribution}
\end{align}
Since the marginalized distribution is only two dimensional, it can easily be estimated by sampling from the learned generative model and subsequently using some simple density estimation technique like kernel density estimation (KDE)~\cite{kde}.\\
Using this metric has two advantages. Firstly, when combined with MoN, it gives a better estimation of how well the model really learned the underlying PDF as the marginalized log likelihood favours a model with sharper probability but ignores inter-time step dependencies. The MoN metric, on the other hand, can give a decent estimate of the joint probability of prediction for all time steps even for large $T$. Secondly, it is useful to have the per time step probability distribution to generate the grid based cost map in order to do ego path planning~\cite{path_planning_cost_map} in autonomous driving. We will elaborate further in section \ref{sec:recover_maximum_likelihood}.


\section{Recovering the ground truth PDF}
\label{sec:reover_gt}
\subsection{Sample from squared distribution}
\begin{figure}
     \begin{subfigure}[b]{0.47\linewidth}
         \centering
         \includegraphics[width=\linewidth]{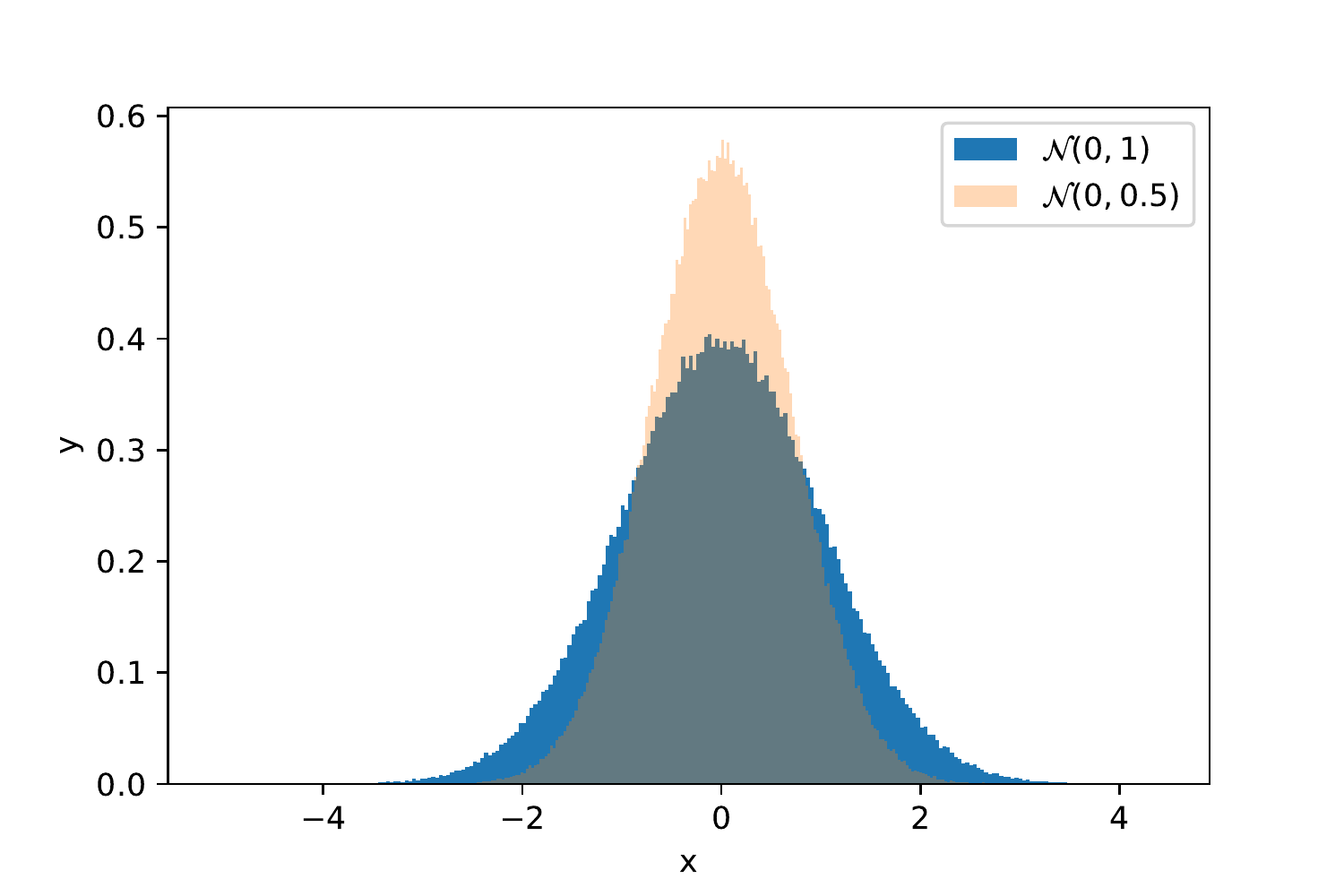}
         \caption{Histogram of the Normal Gaussian (blue) and the analytically squared Gaussian (orange). }
     \end{subfigure}
     \hspace{.1cm}
     \begin{subfigure}[b]{0.47\linewidth}
         \centering
         \includegraphics[width=\linewidth]{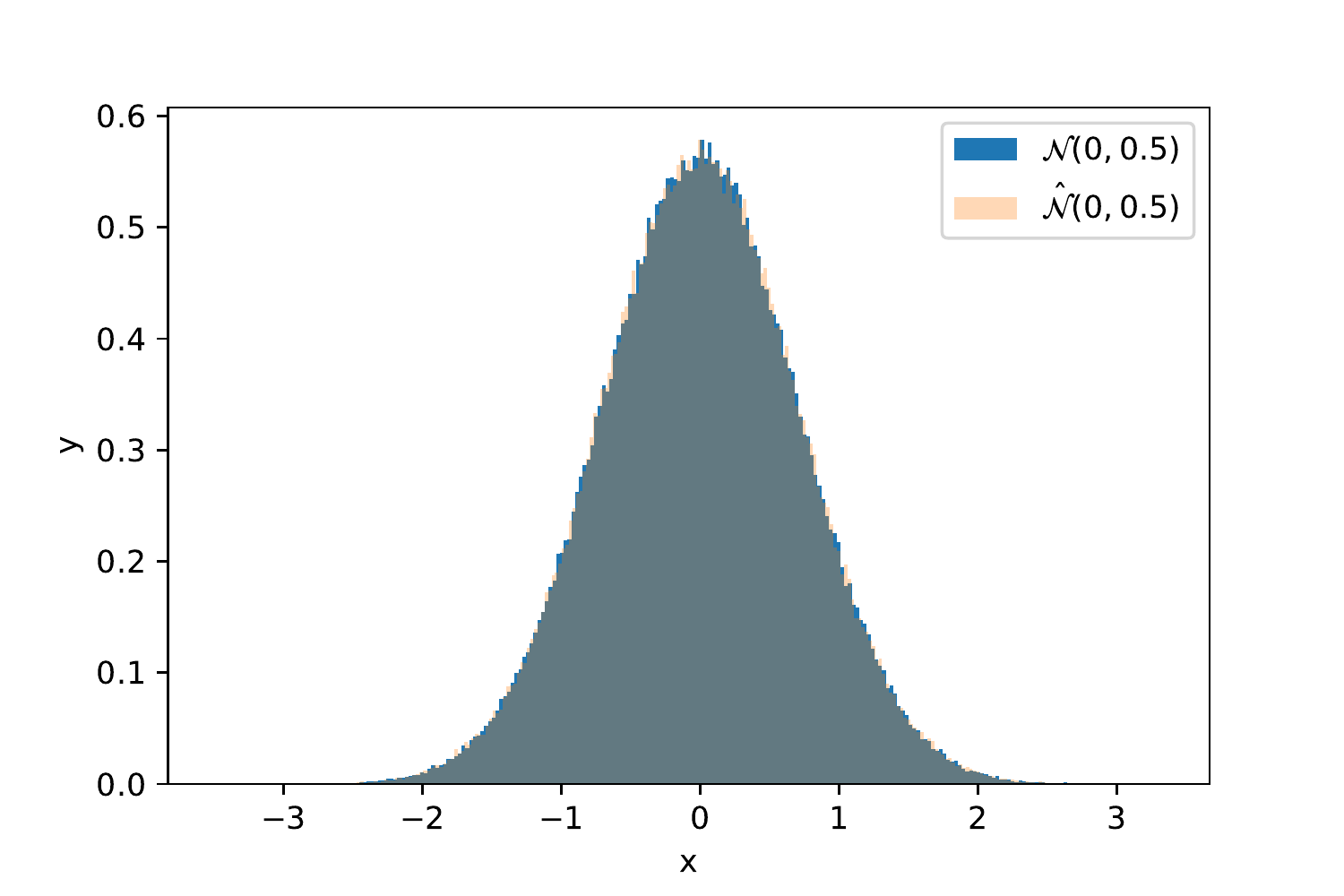}
         \caption{Histogram of the analytically squared Gaussian (blue) and the estimated squared Gaussian (orange).}
     \end{subfigure}
     \caption{(a) Histogram of samples from the Normal Gaussian $\mathcal{N}(0,1)$ and the analytically squared Gaussian $\mathcal{N}(0,\sqrt{0.5})$. (b) Histogram of samples from $\mathcal{N}(0,\sqrt{0.5})$ and samples from  $\Hat{\mathcal{N}}(0,\sqrt{0.5})$ that was obtained by applying the squaring compensation on $\mathcal{N}(0,1)$. It is clearly visible, that $\Hat{\mathcal{N}}(0,\sqrt{0.5})$ matches $\mathcal{N}(0,\sqrt{0.5})$ very closely.}
     \label{fig:square_sampling}
\end{figure}

Assuming a learner $P(\boldsymbol{x})$ converged towards $\sqrt{P_T(\boldsymbol{x})}$, we now want to recover the ground truth PDF $P_T(\boldsymbol{x})$. For low--dimensional tasks, we show here a simple way to sample from $P(\boldsymbol{x})^2$ thereby cancelling the square root. For this consider a PDF $P(\boldsymbol{x})$ and bin it in bins of width $\epsilon$. Then the probability that two iid samples fall in the same bin $b_i$ is 
\begin{align}
    \int_{b_i} \int_{b_i} P(\boldsymbol{x_1}, \boldsymbol{x_2}) \diff \boldsymbol{x_1} \diff \boldsymbol{x_2} &= \int_{b_i} \int_{b_i} P(\boldsymbol{x_1}) \cdot P(\boldsymbol{x_2}) \diff \boldsymbol{x_1} \diff \boldsymbol{x_2} \notag\\
    &= \int_{b_i} P(\boldsymbol{x_1}) \diff \boldsymbol{x_1} \cdot \int_{b_i} P(\boldsymbol{x_2}) \diff \boldsymbol{x_2}  \notag\\
    &= P(b_i) \cdot P(b_i) = P(b_i)^2 
\end{align}
This can be realized by two different algorithms:
\begin{itemize}  
\item Bin the sample space in bins of width $\epsilon$. Sample from $P(\boldsymbol{x})$, and count in which bin the sample falls. Repeat this until there are two samples in one bin, and choose one of those samples.
\item Sample two times from $P(\boldsymbol{x})$. If the samples have a distance of less then $|\boldsymbol{x_1} - \boldsymbol{x_2}| < \epsilon$, choose one of those samples. Otherwise repeat. 
\end{itemize}
The two variations are a trade-off of speed vs memory, with the first one being faster but more memory intensive. Primarily, these two can be used during inference time, to generate samples that come from the target distribution. Without testing this, for unconditional problems one could also think of using it during training with stochastic gradient descent~\cite{stochastic_gradient_descent} to sample the data point shown to the network. However, due to the curse of dimensionality, this is only possible in relatively low dimensions (or in higher dimensions, if the distribution has a very small support), since otherwise it is too unlikely to generate two samples with $|\boldsymbol{x_1} - \boldsymbol{x_2}| < \epsilon$ for small enough $\epsilon$.\\
To validate the algorithm, we sampled from $\mathcal{N}(0,1)$ and from $\frac{1}{C}\mathcal{N}^2(0,1) = \mathcal{N}(0,\sqrt{0.5})$ and with the proposed algorithm from $\mathcal{N}(0,1)$ which gives us an estimate of $\frac{1}{C}\mathcal{N}^2(0,1)$ that we denote as $\Hat{\mathcal{N}}(0,\sqrt{0.5})$. The results are depicted in Figure \ref{fig:square_sampling}.
\subsection{Maximum likelihood based recovery}
\label{sec:recover_maximum_likelihood}
In the previous section, we assumed that the learner converged to $\sqrt{P_T(\boldsymbol{x})}$. However, there are several reasons, why this might not be exactly the case: the model is not expressive enough, the training got stuck in a local minima, $N$ was too small or MoN was used in addition to other losses, like the adversarial loss, that does converge to $P_T(\boldsymbol{x})$. In those cases, squaring the learned distribution could actually move it farther away from the ground truth distribution. At least for some very specific applications, there is still a possibility to compensate for the dilating effect of MoN. One of these applications is trajectory prediction for path planning in autonomous cars. A possible approach to path planning is to create a cost map~\cite{path_planning_cost_map}, and then run a path finding algorithm that minimizes these costs under the physical constrains of the vehicle dynamics. In this framework, one can imagine that the probability of a traffic participant being at a certain point in time can simply be framed as costs on the cost map. Finding a path that minimizes these costs then is equivalent to minimizing the probability of crashing with another traffic participant. Since these algorithms only care about whether there will likely be a traffic participant at a certain point in time at a certain point in space, and not how it got there, we only care about the marginalized probability distribution per time step. Since this distribution is only two dimensional, we can easily estimate it by sampling from the trained model and using a kernel density estimator~\cite{kde} to recover the PDF. The bandwidth of the KDE can be selected via cross validation on a left out set of generated samples \cite{bandwidth_selection}. Subsequently, the KDE can be evaluated on a grid and a transformation as defined in \eqref{eq:transformation} can be applied for various $\Bar{k}$. At the end the $\Bar{k}$ is selected, that maximizes the log likelihood of the ground truth sample for each of these samples. Algorithm \ref{algorithm:max_likelihood} makes these steps precise.
\begin{algorithm}
    \caption{Algorithm to find the best compensation parameter $\Bar{k}$ for transformation $T_{\Bar{k}}(P)$ of model $P(\boldsymbol{X}|\boldsymbol{I})$ under the inputs \{$\boldsymbol{I}_i\}_{i=0,...,K}$ so that the marginalized log likelihood of the ground truth sample $\{(x^*_i,y^*_i)\}_{i=0,...,K}$ is maximized. Here, $\{\Bar{k}\}_\text{search}$ is the search space and $(\boldsymbol{x}, \boldsymbol{y})$ are grid points. $n_\text{sample}$ is an sufficiently big integer and  $\alpha_\text{split} \in (0,1)$.}
    \label{algorithm:max_likelihood}
    \begin{algorithmic}[1] 
        \Procedure{FindBestCompensationParameter}{\hspace{-0.11cm}$P(\boldsymbol{X}|\boldsymbol{I})$, \{$\boldsymbol{I}_i\}_{i=0,...,K}$, $\{(x^*_i,y^*_i)\}_{i=0,...,K}$, $n_\text{sample}$, $\alpha_\text{split}$, $\{\Bar{k}\}_\text{search}$, $(\boldsymbol{x}, \boldsymbol{y})$} 
            \State $L_\text{max} \leftarrow -\infty$ 
            \State $\Bar{k}_\text{best} \leftarrow 0$
            \For{$\Bar{k}$ in $\{\Bar{k}\}_\text{search}$}
                \State $L_\text{run} \leftarrow 0$ 
                \For{$\boldsymbol{I}_i$ in \{$\boldsymbol{I}_i\}_{i=0,...,K}$}
                    \State $\{s_i^j\}_{j=0,...,n_\text{sample}} \stackrel{\text{iid}}{\sim} P(\boldsymbol{\boldsymbol{X}_t}|\boldsymbol{I}_i)$ 
                    \State Use $\{s_i^j\}_{j=0,...,\alpha_\text{split}n_\text{sample}}$ to fit a KDE and $\{s_i^j\}_{j=\alpha_\text{split}n_\text{sample},...,n_\text{sample}}$ to find best bandwidth for the KDE \cite{bandwidth_selection}
                    \State $L(\boldsymbol{x}, \boldsymbol{y}) \leftarrow$ evaluate KDE on grid $(\boldsymbol{x}, \boldsymbol{y})$
                    \State $L(\boldsymbol{x}, \boldsymbol{y}) \leftarrow \frac{L(\boldsymbol{x}, \boldsymbol{y})^{\Bar{k}} } {\sum_{\boldsymbol{x}, \boldsymbol{y}} L(\boldsymbol{x}, \boldsymbol{y})^{\Bar{k}} }$
                    \State $L_\text{run} \leftarrow L_\text{run} + \log L(x^*_i, y^*_i)$
                \EndFor
                \If{$L_\text{run} > L_\text{max}$}
                    \State $L_\text{max} \leftarrow L_\text{run}$
                    \State $\Bar{k}_\text{best} \leftarrow \Bar{k}$
                \EndIf
            \EndFor \\
        \Return{$\Bar{k}_\text{best}$}
        \EndProcedure
    \end{algorithmic}
\end{algorithm}
The parameter $\Bar{k}_\text{opt}$ can then be found to improve the estimated PDF during inference time by doing the steps of the innermost loop in algorithm \ref{algorithm:max_likelihood} with the found $\Bar{k}$. Note that this compensation is very lightweight, once $\Bar{k}_\text{opt}$ is found, for tasks where the KDE reconstruction has to be done anyway.


\section{Experiments}
\label{sec:experiments_toy}
\subsection{MoN minimum of Mixture of Gaussians}
\label{sec:experiments_toy_1}
\begin{figure}
     \centering
    \includegraphics[width=\linewidth]{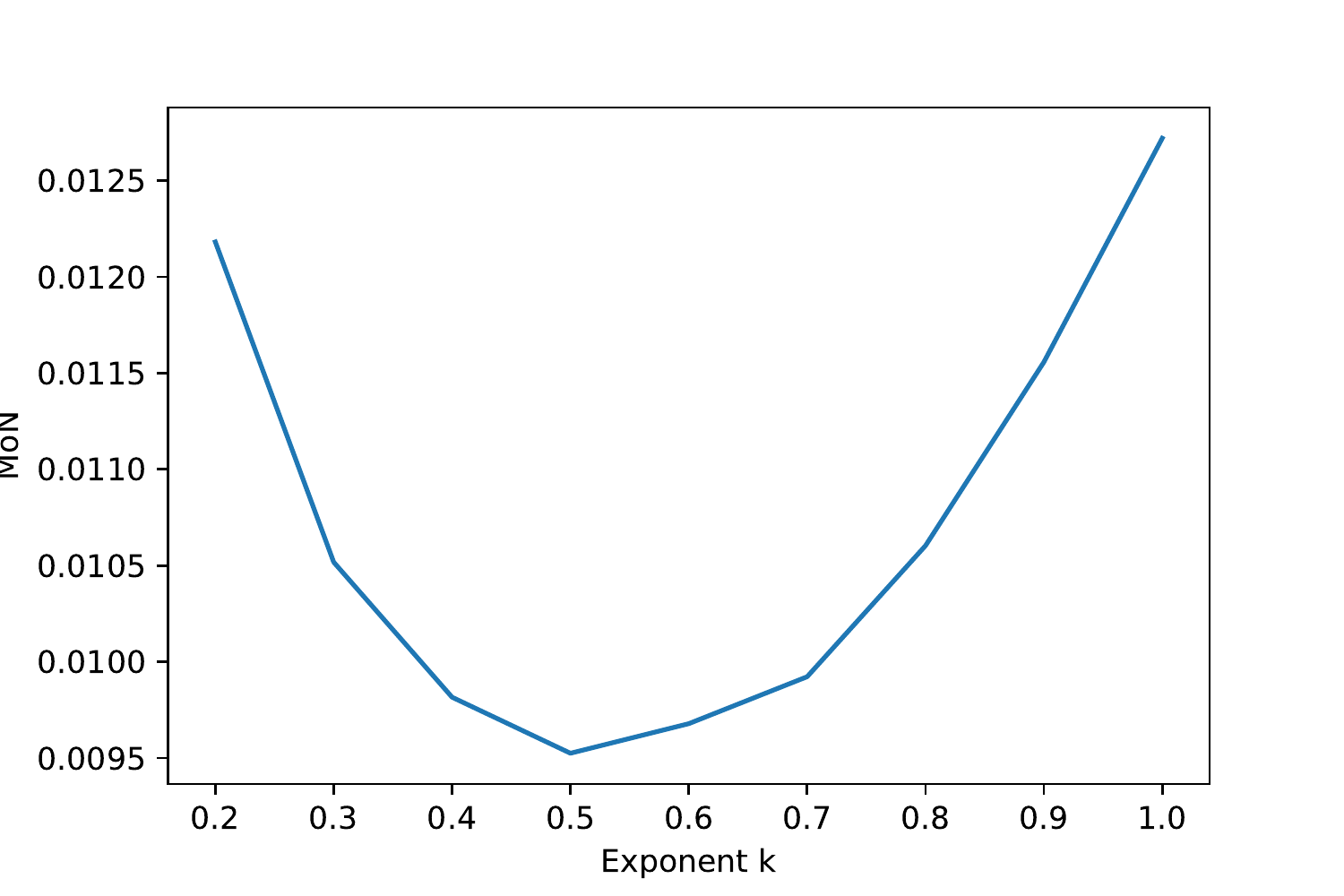}
    \caption{MoN values for different $k$ by using samples from $f_1$ as the groundtruth distribution and $f_1^k$ as the test distribution with $N=256$}
    \label{fig:exponent_vs_mon_mixture_normal}
\end{figure}

We verify our result on two toy experiments: \\
For the first experiment we sample $M=50000$ times from
\begin{align}
    \{x_{1,i}\}_{i=1,...,M} \stackrel{\text{iid}}{\sim} f_1 := \mathcal{N}(0,1)  
\end{align}
\/*
and from 
\begin{align}
    \{x_{1,i}\}_{i=1,...,N} \stackrel{\text{iid}}{\sim} f_1 := \frac{1}{C_{f_1}} \left(\mathcal{N}(0,\sqrt{3/2}) + \mathcal{N}(4,1)\right)
\end{align}
with the appropriate normalization constant $C_{f_2}$.\\
*/
We then consider the family of PDFs
\begin{align}
    \{x_{1,i}^k\}_{i=1,...,256} \stackrel{\text{iid}}{\sim} f_1^k := \frac{1}{C_{{f_1},k}} \mathcal{N}(0,1)^k
\end{align}
\/*
and 
\begin{align}
    \{x_{1,i}^k\}_{i=1,...,128} \stackrel{\text{iid}}{\sim} f_1^k := \frac{1}{C_{{f_1},k}}\left(\mathcal{N}(0,\sqrt{3/2}) + \mathcal{N}(4,1)\right)^k
\end{align}
*/
and sample 256 data points. Subsequently we calculate for each of the $M$ sample from the original distribution the minimum distance from the 256 samples. All of this is averaged over $R=100$ tries. Note, that we do not learn a model here, but merely search for the $k$, that minimizes the MoN loss for the respective PDFs. The results are reported in Figure \ref{fig:exponent_vs_mon_mixture_normal}.
As expected, the minimum value is within our search resolution exactly $k=0.5$, which means the PDF that minimizes the MoN loss is the square root of the ground truth PDF. This validates Theorem \ref{theorem}.

\subsection{Learn Mixture of Gaussians}
\begin{figure}
\begin{subfigure}{.47\linewidth}
   \centering
    \includegraphics[width=\linewidth]{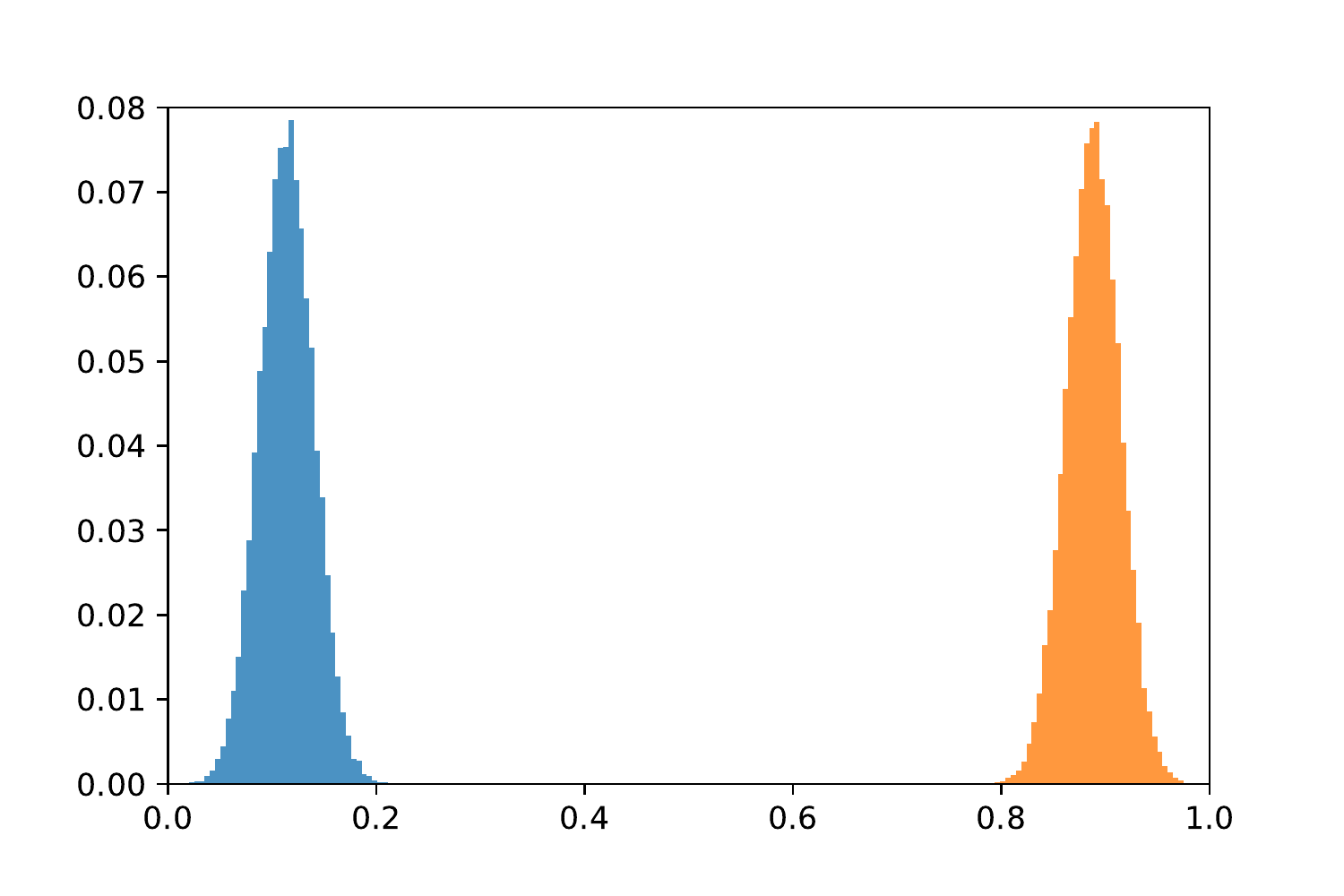}
    \caption{Color coded input samples from \eqref{mon_input_1} (blue) and \eqref{mon_input_2} (orange)}
    \label{fig:mon_input}
\end{subfigure}
\hspace{.1cm}
\begin{subfigure}{.47\linewidth}
  \centering
  \includegraphics[width=\linewidth]{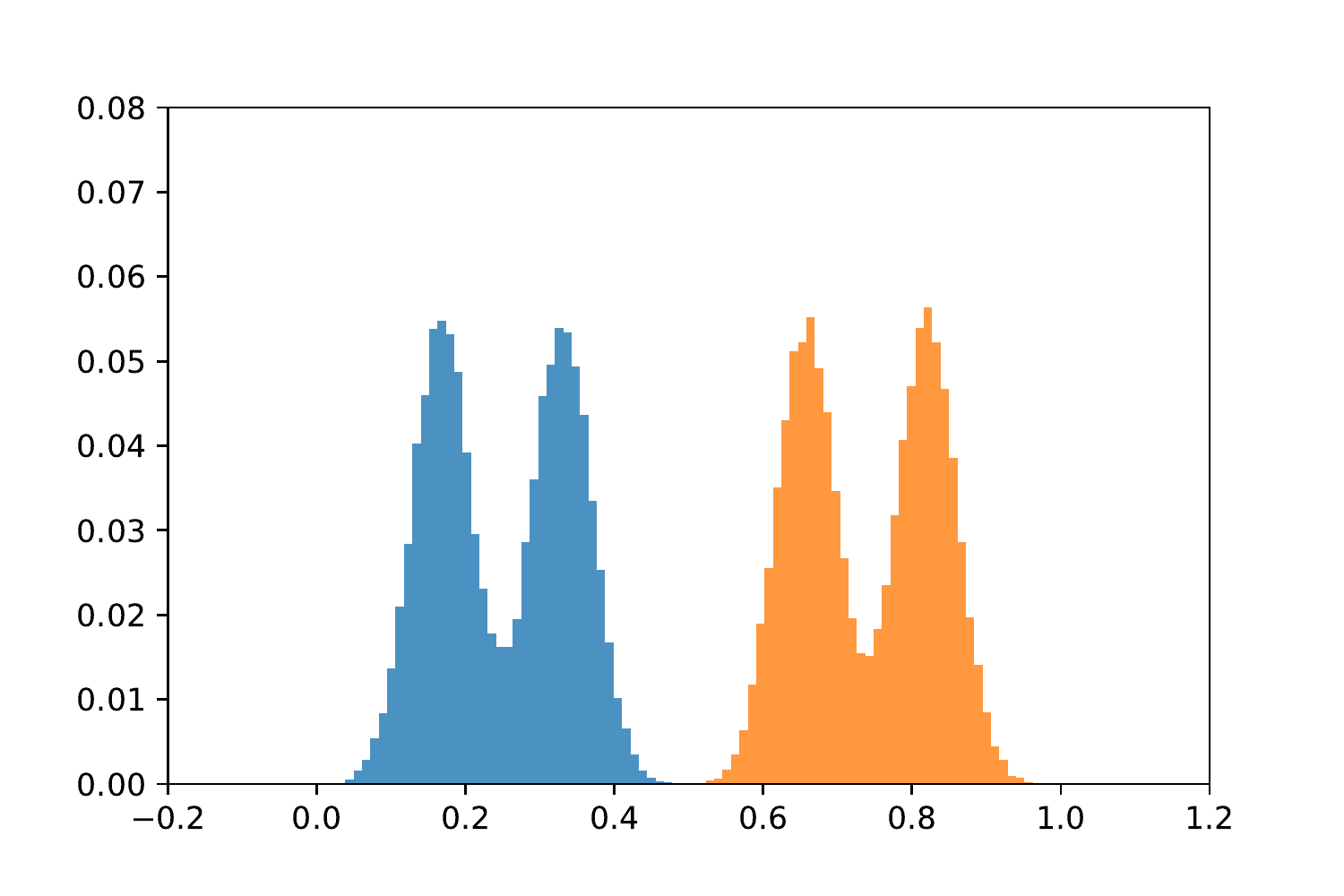}
  \caption{Color coded target samples from \eqref{mon_target_1} (blue) and \eqref{mon_target_2} (orange)}
  \label{fig:mon_target}
\end{subfigure}
\begin{subfigure}{.47\linewidth}
  \centering
  \includegraphics[width=\linewidth]{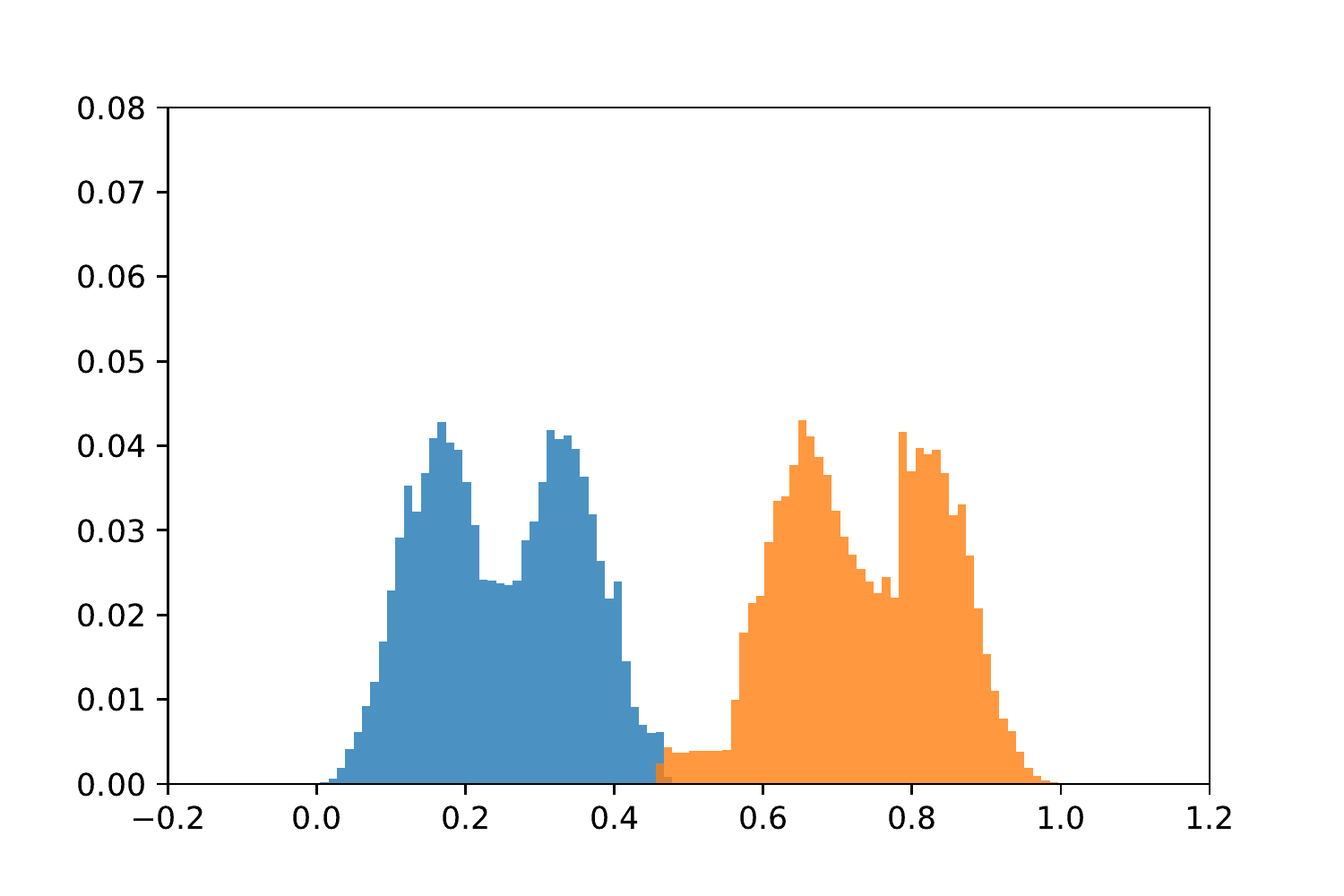}
  \caption{The learned distribution $P_\text{learned}$}
  \label{fig:mon_learned}
\end{subfigure}
\hspace{.1cm}
\begin{subfigure}{.47\linewidth}
  \centering
  \includegraphics[width=\linewidth]{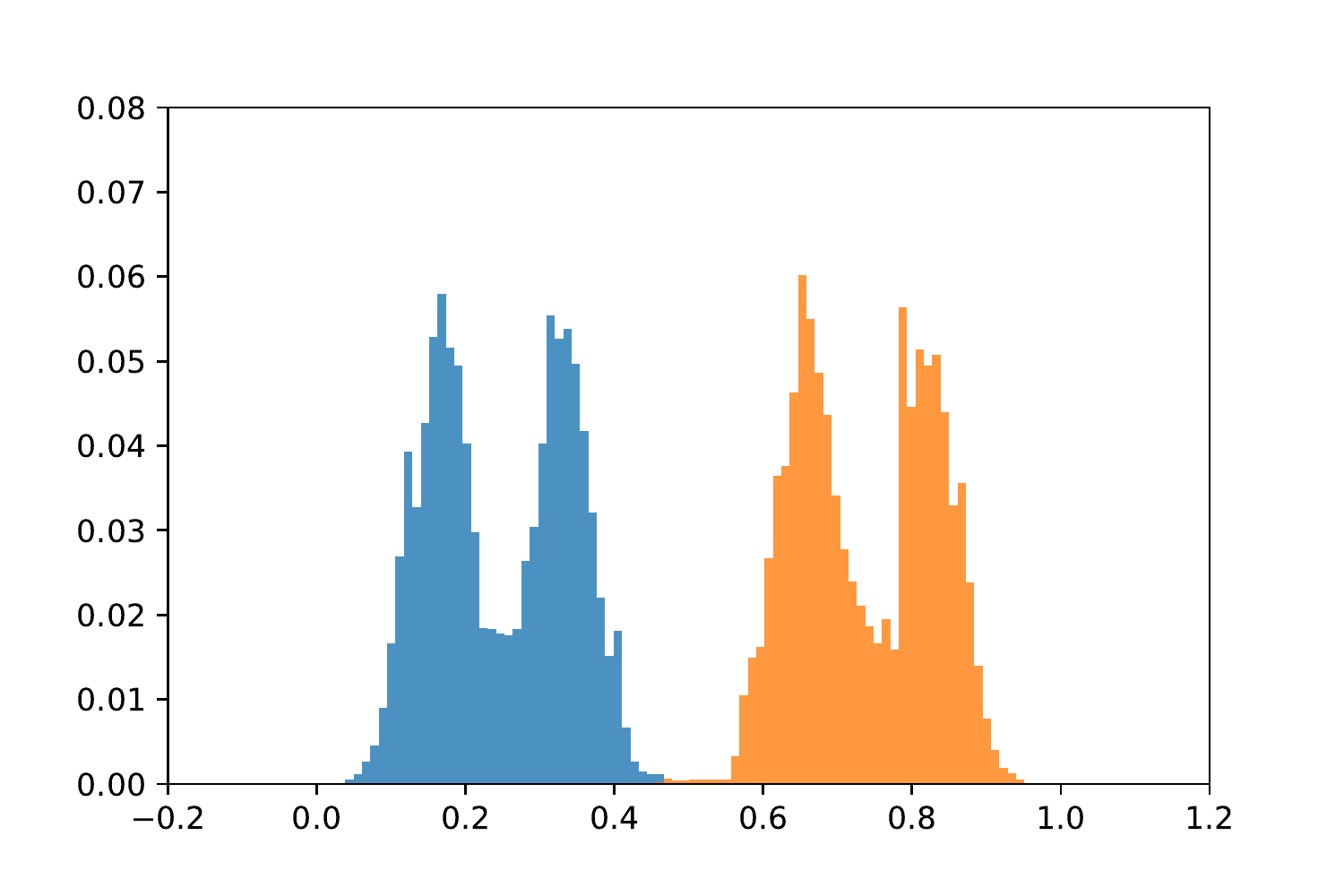}
  \caption{The learned distribution squared $P_\text{learned}^2$}
  \label{fig:mon_compensated}
\end{subfigure}
\caption{The input distribution and corresponding targets distribution are shown in (a) and (b) respectively. In (c), samples from the learned distribution $P_\text{learned}$ are plotted and (d) shows $P_\text{learned}^2$ which is square of the learned distribution. Obviously $P_\text{learned}^2$ matches the ground truth distribution better.}
\label{fig:fig}
\end{figure}

Theorem \ref{theorem} and the previous experiment show that the PDF that minimizes the MoN loss for big N is actually the square root of the ground truth PDF. It is not clear though, if a generative model, trained with the MoN loss actually converges towards this solution or if it gets stuck in local minima. We test this with another toy dataset and a very simplistic generative model: the dataset consist of inputs, which are randomly sampled either from 
\begin{align}
    f_2 = \mathcal{N}(-1.5,0.1)  \label{mon_input_1}
\end{align}
or 
\begin{align}
    f_3 = \mathcal{N}(+1.5,0.1)  \label{mon_input_2}
\end{align}
and of targets, which are randomly sampled from 
\begin{align}
    f_{2, \text{target}} = \frac{1}{C_{{f_2}, \text{target}}}\left(\mathcal{N}(-2,1) + \mathcal{N}(-4,1)\right)  \label{mon_target_1}
\end{align}
or 
\begin{align}
    f_{3, \text{target}} = \frac{1}{C_{{f_3}, \text{target}}}\left(\mathcal{N}(2,1) + \mathcal{N}(4,1)\right)  \label{mon_target_2}
\end{align}
respectively. The inputs and targets are illustrated and color coded in Figure \ref{fig:mon_input} and \ref{fig:mon_target}. \\
For the generative model, we used a very simple neural network consisting of an encoder that predicts the mean and variance of a Gaussian by encoding the samples from distribution \ref{mon_input_1} or \ref{mon_input_2} to predict parameter of a Gaussian, and a decoder that takes $N$ samples from the Gaussian and transforms them to minimize the MoN loss (for architecture details we refer to the supplementary details). Note that the model would not be able to learn the correct distribution with a simple mean squared error loss, as it would only learn to generate the mean of the distribution.\\
We train the model with the MoN loss with $N=128$. However we noticed that this consistently led to poor local minima, where the modes that are farther away from the center were poorly predicted. We found it vastly helpful to start with a low $N$, and then slowly increase it during training till the final $N$ is reached. The resulting learned PDF is depicted in Figure \ref{fig:mon_learned}. As one can see, the learned PDF looks dilated. However, since theorem \ref{theorem} tells us that this should be approximately the square root of the ground truth PDF, we can simply square and normalize over the bins, to recover the ground truth. This is shown in \ref{fig:mon_compensated}. The qualitative superiority of the compensated PDF is obvious. Also the numerically estimated Jensen--Shannon divergence becomes almost an order of magnitude smaller (See Table \ref{sjd_table}).
\renewcommand{\arraystretch}{1.2}
\begin{table}
\centering
\begin{tabular}{|l|l|l|}
\hline 
Metric & $P_\text{learned}$ & $P_\text{learned}^2$ \\ \hline \hline
\text{\text{Jensen--Shannon divergence}} & 0.1282  & 0.0191 \\ \hline 
\end{tabular}  
\caption{The JS divergence between the ground truth PDF and the learned PDF $P_\text{learned}$ is worse than that between the ground truth and the compensated version $P_\text{learned}^2$.}
\label{sjd_table}
\end{table}
\subsection{Dependence of minimizing exponent on N}
\begin{figure}
    \centering
    \includegraphics[width=.99\linewidth]{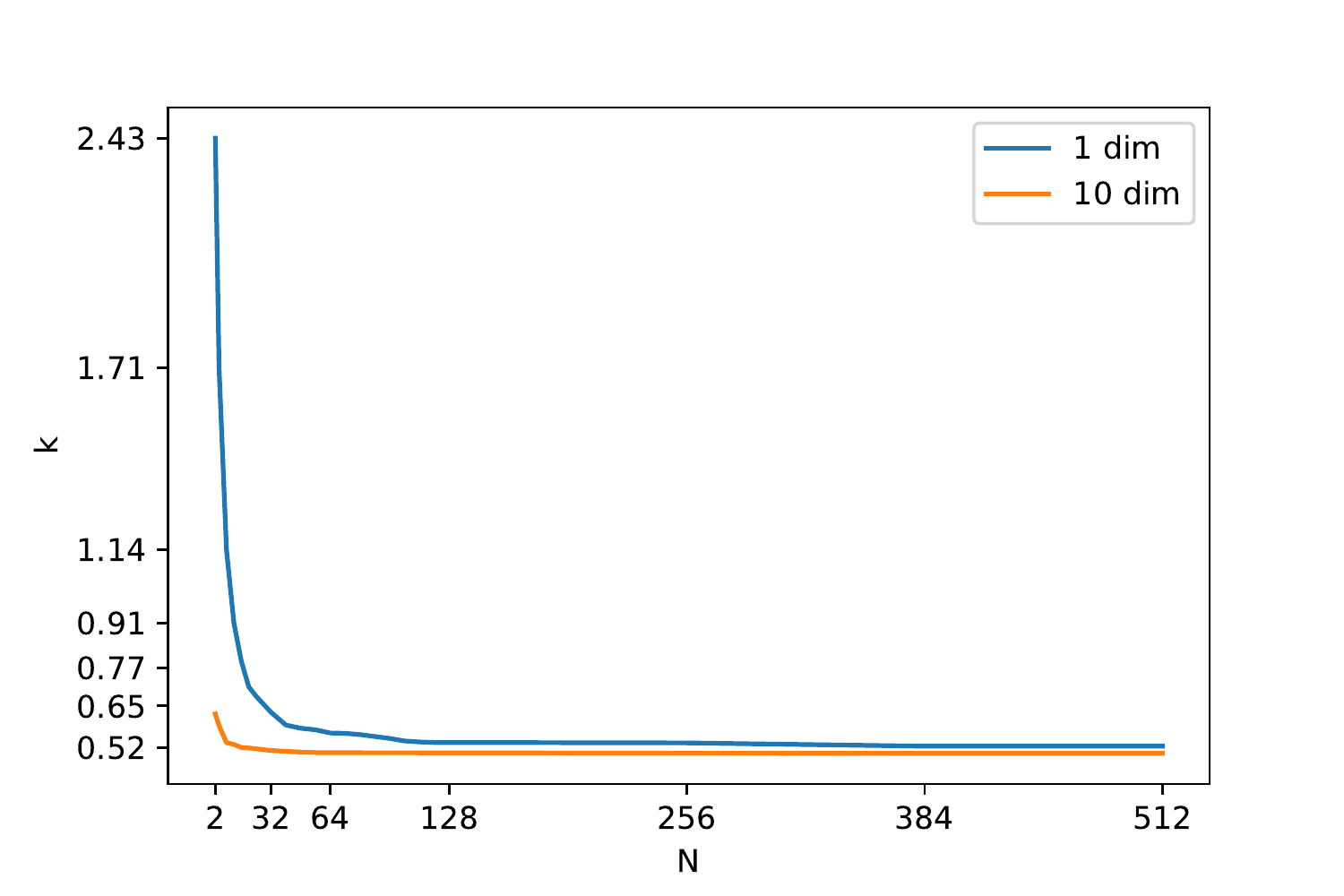}
    \caption{The variation of the $k$ that minimizes the MoN loss is plotted with respect to $N$ for PDFs with dimensionality 1 and 10. Note that the 10 dimensional one converges much faster. This implies that a widespread PDF is preferred by MoN in higher dimensions even for small $N$.}
    \label{fig:minimizing_exponent_vs_N_both}
\end{figure}

Next we want to verify proposition \ref{proposition} experimentally, by repeating the experiment from section \ref{sec:experiments_toy_1}. This time however we search for the MoN minimizing exponent $k$ for different $N$. The results are plotted in Figure \ref{fig:minimizing_exponent_vs_N_both}. It is obvious that Proposition \ref{proposition} holds at least for this particular PDF. The same experiment is repeated with a 10 dimensional version of the PDF (See Figure \ref{fig:minimizing_exponent_vs_N_both}). Surprisingly, the results imply that for higher dimensional PDF, the MON loss prefers $k$ close to 0.5 even for small $N$. This is especially important in the context of using MoN as a metric.


\section{Application to trajectory prediction in autonomous vehicles}
The problem considered here is to find a model $\boldsymbol{P(Y|\text{I})}$, where $\boldsymbol{Y}$ is a trajectory with $\boldsymbol{Y} = \{ (x_1, y_1), (x_2, y_2), ..., (x_T, y_T)\}$ of length $T$ and $\text{I}$ is the input. We experiment with the prediction of highway vehicles and pedestrian trajectories.
\subsection{NGSIM Dataset}
\label{sec:ngsim}
In this section, we will train a generative model using MoN on the Next Generation Simulation (NGSIM) dataset and show that compensating the learned probability distribution using Algorithm \ref{algorithm:max_likelihood} will improve the average log likelihood of ground truth samples.\\

\begin{figure}
    \begin{subfigure}{1\linewidth}
        \centering
        \includegraphics[width=.8\linewidth]{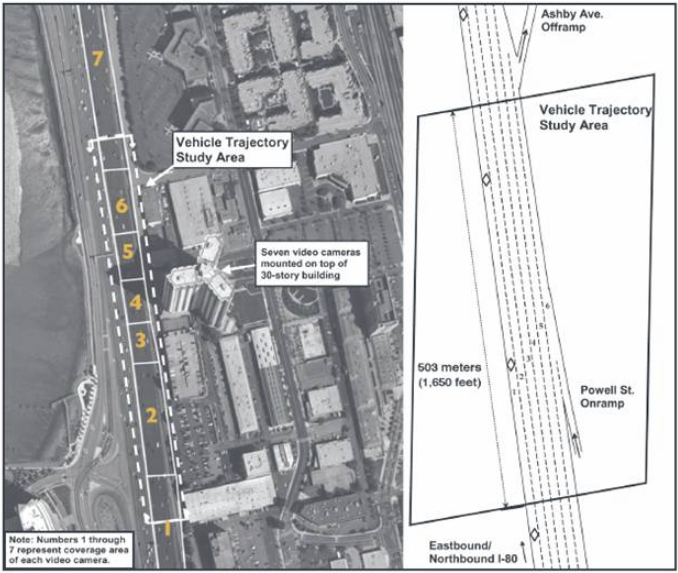}
        \caption{Overview of the NGSIM dataset \cite{ngsim_dataset}}
        \label{fig:ngsim}
    \end{subfigure}\
    \begin{subfigure}{1\linewidth}
        \centering
        \includegraphics[width=.8\linewidth]{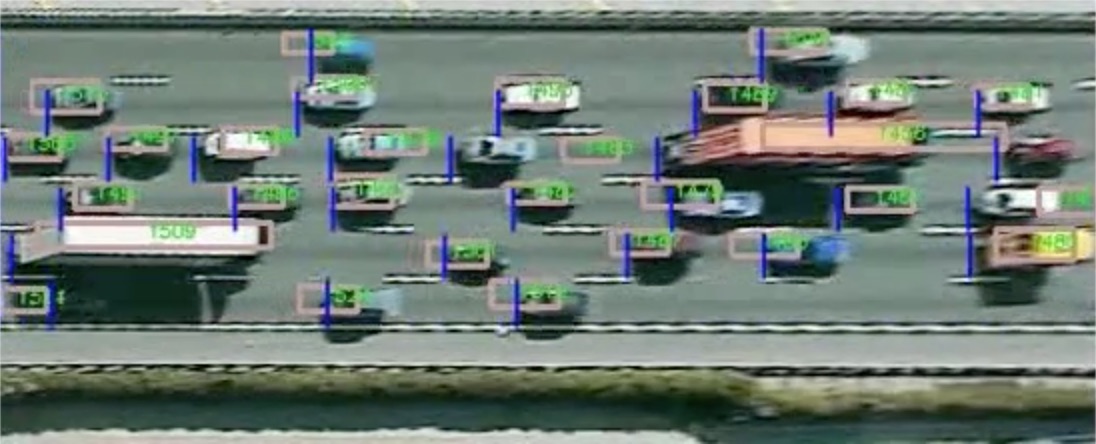}
        \caption{Close up of the highway and tracking of the vehicle \cite{ngsim_dataset_close_up}.}
        \label{fig:ngsim_close}
    \end{subfigure}
    \caption{(a) An overview of the highway section the NGSIM dataset was recorded on~\cite{ngsim_dataset}. (b) Close up of the I-80 with a visualization of the tracking the vehicle.}
    \label{fig:i80}
\end{figure}

\noindent The NGSIM dataset consists of 45 minutes of vehicles tracked along a section of the I-80 highway which is approximately 0.5 kilometer long (see Fig \ref{fig:i80}).\\
Our generative model consists of an LSTM with 128 units that encodes the trajectory of a vehicle and predicts the parameters of a 12 dimensional Gaussian distribution. Then we sample $N=100$ times from this distribution. These samples are encoded by 2 dense layers, each with 128 units and ReLu activations. Finally, a decoder LSTM with 128 units predicts the $\Delta x_t$ and $\Delta y_t$, so that $x_{t-1} + \Delta x_t = x_t$ and $y_{t-1} + \Delta y_t = y_t$ respectively. We downsample the data by a factor of 16 and consider 3 time steps, which amounts to a time horizon of 4.8 seconds. Since the vehicle moves much faster in x direction than in y direction, which means the errors in x direction are much higher, we weight the error in y direction with a factor of 20 during training (not during test time).\\

\begin{figure}
    \centering
    \includegraphics[width=.99\linewidth]{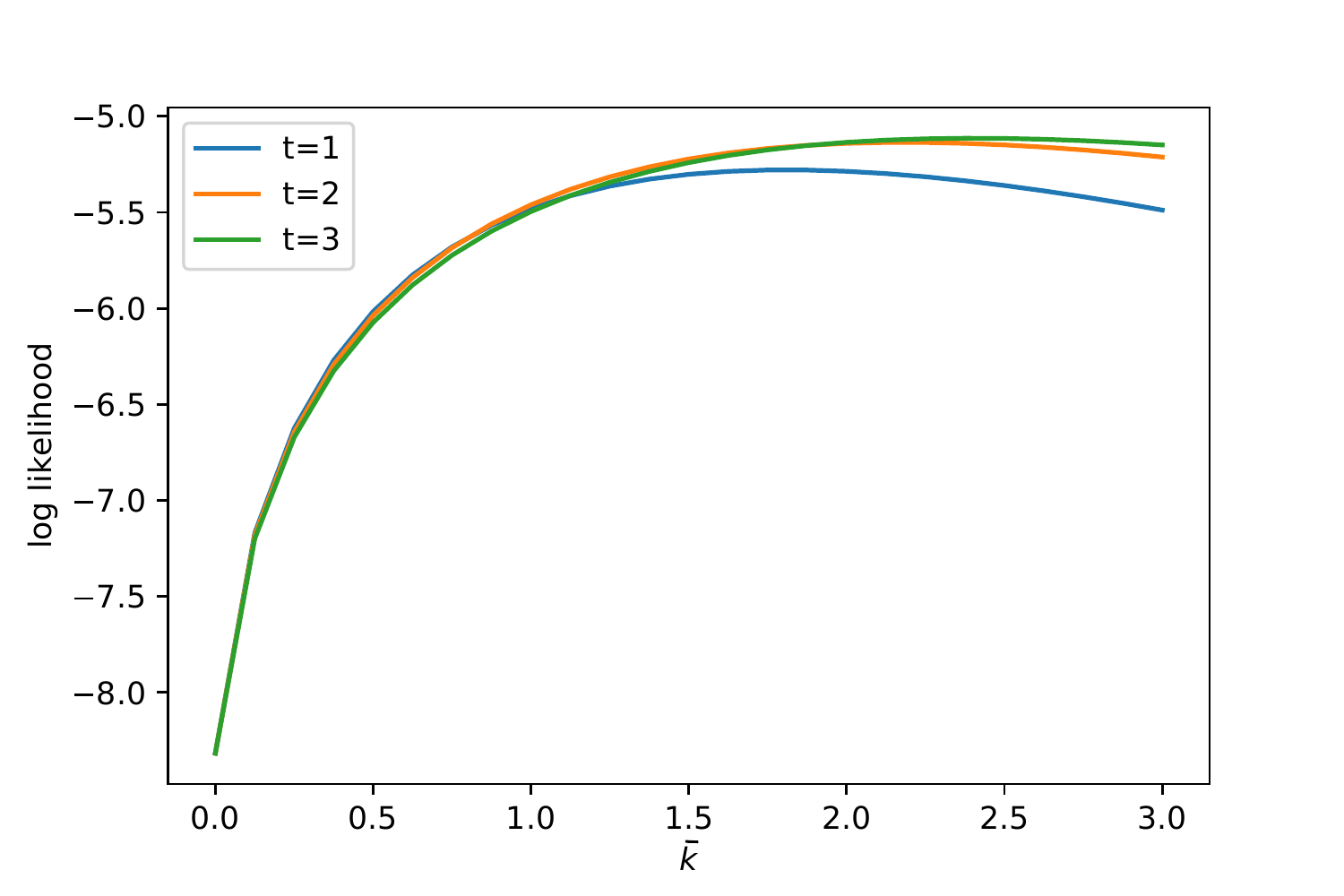}
    \caption{The average log likelihood in dependency of the $\Bar{k}$ used for the transformation in \eqref{eq:transformation}.}
    \label{fig:likelihood_k_ngsim}
\end{figure}

As described in \ref{sec:recover_maximum_likelihood}, for the problem of trajectory prediction in the context of path planning with a cost map, it is enough to only consider the marginalized distribution (See the supplementary materials for plots of the uncompensated marginalized PDF, reconstructed with a KDE as described in \ref{sec:recover_maximum_likelihood}).\\
Since we are using the MoN loss, the learned PDF has to be compensated for the dilation effect. We apply algorithm \ref{algorithm:max_likelihood} for this purpose. We set $n_\text{sample}$ to 1000 and $\alpha_\text{split}$ to 0.7. As the set of possible compensation parameter $\{\Bar{k}_\text{search}\}$, we use 25 values between 0.001 and 3. 
Our experiments showed $\Bar{k}_\text{opt,t=1} = 1.88$, $\Bar{k}_\text{opt,t=2} = 2.12$ and $\Bar{k}_\text{opt,t=3}=2.12$ which are close to the expected value of $2$. A plot of the average log likelihood dependency from the chosen $\Bar{k}$ is shown in Figure \ref{fig:likelihood_k_ngsim}. The supplementary material shows the compensated reconstructed PDFs.
Furthermore, if we use Algorithm \ref{algorithm:max_likelihood} to find the $\Bar{k}_\text{opt}$ that is optimal for all 3 time steps simultaneously, the algorithm yields $\Bar{k}_\text{opt} = 2.00$ withing the search resolution, which is exactly the theoretically expected value. 
After obtaining the $\Bar{k}_\text{opt}$, we applied the compensation on a left out test dataset and observed an improvement in the average log likelihood of ground truth trajectories. The results (see table \ref{results:ngsim_ll}) show that our compensated PDF clearly outperforms the uncompensated one.

\begin{table}[]
\begin{tabular}{|l|l|l|l|}
\hline
PDF               & $\Lagr^1_{\textbf{D}_\text{Test}}(P)$ & $\Lagr^2_{\textbf{D}_\text{Test}}(P)$ & $\Lagr^3_{\textbf{D}_\text{Test}}(P)$ \\ 
\hline
\hline
Original PDF & -5.48 & -5.46 & -5.50  \\
\hline
Compensated PDF & \textbf{-5.28} & \textbf{-5.13} & \textbf{-5.12} \\
\hline
\end{tabular}
\caption{The results for the marginalized log likelihood as defined in \eqref{eq:marginalized_distribution} on the NGSIM dataset for the first 3 time steps (4.8 seconds). The compensated PDF consistently outperforms the uncompensated one.}
\label{results:ngsim_ll}
\end{table}

\subsection{Social--GAN}
\label{sec:gan_experiments}
\begin{figure}
    \centering
    \includegraphics[width=.99\linewidth]{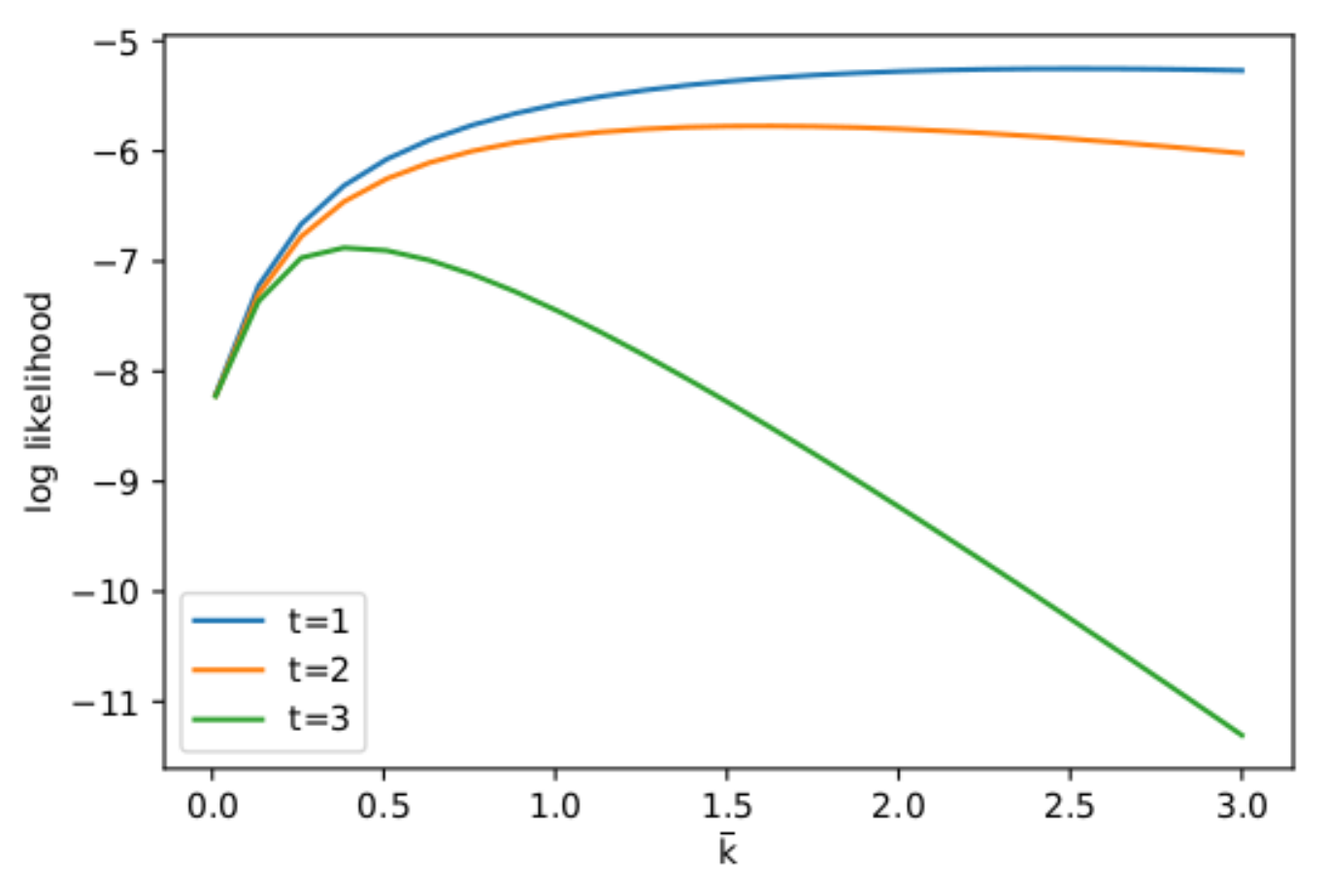}
    \caption{The average log likelihood in dependence of the $\Bar{k}$ used for the transformation in
    \eqref{eq:transformation}.}
    \label{fig:likelihood_k}
\end{figure}

Next, we experiment on pedestrian trajectory data with Social--GAN~\cite{social_gan} to show that even a state--of--the--art model can be improved by using our proposed compensation. Here, the authors use a combination of the MoN loss and the adversarial loss. They also designed a social pooling mechanism for efficient modelling of the social interactions of the pedestrians. We consider the Zara 1 dataset \cite{zara_dataset} and use the best performing model provided by the authors of \cite{social_gan} (\href{<https://github.com/agrimgupta92/sgan>}{https://github.com/agrimgupta92/sgan}).
The Zara dataset consists of 489 trajectories extracted from 13 minutes of videos on the corner of a sidewalk in a city (See Figure \ref{fig:zara}). In the supplementary materials, a few plots of the uncompensated PDFs are shown.
\begin{figure}
    \centering
    \includegraphics[width=.8\linewidth]{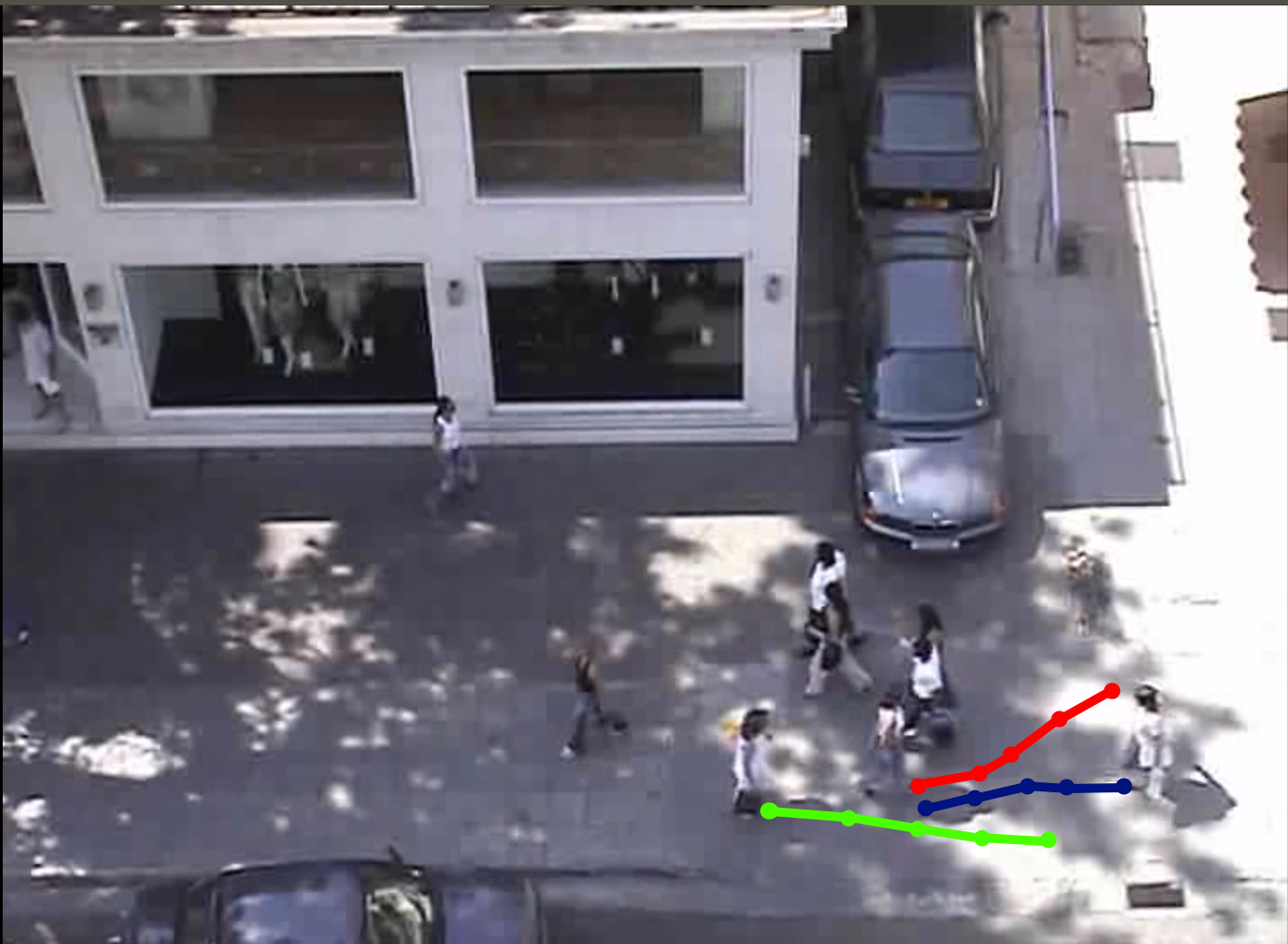}
    \caption{Illustration of the Zara dataset with ground truth trajectories.}
    \label{fig:zara}
\end{figure}
We use algorithm \ref{algorithm:max_likelihood} with the same settings as in \ref{sec:ngsim}. The resulting optimal compensation parameters are $\Bar{k}_\text{opt,t = 1} = 2.50$ and $\Bar{k}_\text{opt,2}=1.63$. The variation of the average log likelihood with respect to the chosen $\Bar{k}$ is shown in Figure \ref{fig:likelihood_k}. The supplementary materials show the compensated reconstructed PDFs. The final results of the marginalized log likelihoods are presented in Table \ref{results:social_gan} where we clearly see an advantage over the uncompensated version. For $t \geq 3$ and more difficult pedestrian datasets, this compensation however does not work. This is probably because too many samples fall in the low probability regions of the PDF that was learned by Social--GAN. Therefore, sharpening the learned distribution moves it even farther away from the ground truth distribution.

\begin{table}[]
\centering
\begin{tabular}{|l|l|l|}
\hline
PDF               & $\Lagr^1_{\textbf{D}_\text{Test}}(P)$ & $\Lagr^2_{\textbf{D}_\text{Test}}(P)$  \\ 
\hline
\hline
Original PDF & -5.57 & -5.87   \\
\hline
Compensated PDF & \textbf{-5.24} & \textbf{-5.77}  \\
\hline
\end{tabular}
\caption{The results for the marginalized log likelihood as defined in \eqref{eq:marginalized_distribution} on the Zara1 dataset for the first 2 time steps. The compensated PDF outperforms the uncompensated one. For $t=3$ the compensation parameter $\Bar{k}_\text{opt}$ is however smaller than 1, which means that it has not learned the PDF well enough and our compensation does not make sense.}
\label{results:social_gan}
\end{table}

\section{Conclusion}
In this paper, we proved that the minimum of the MoN loss is not the ground truth PDF, but instead its square root. We validated this result using different experiments on toy and real world datasets. This means that the PDF that minimizes the MoN is a dilated version of the true one. Restricted to a certain class of PDFs, we also showed empirically that the MoN minimizing PDF becomes monotonically further stretched out with bigger N. This leads us to the conclusion that MoN should not be trusted as the only metric to compare models. For trajectory prediction, we instead advocate to also use the log likelihood of the marginalized PDF. Furthermore, we verify empirically that a learner trained with MoN loss can indeed converge to the square root of the PDF. Finally, we show that for certain low--dimensional applications, it is possible to compensate for the dilating effect of MoN and show that the ground truth dataset is more likely in the compensated distribution. 


{\small
\bibliographystyle{ieee}
\bibliography{egbib}
}

\clearpage
\section{Supplementary Material}
Proof of Theorem 1

\begin{proof}
    For the sake of simplicity we only consider the one dimensional case.\\
    First we bin the support of $P_T$ in $M$ equally sized bins $b_1, b_2,...,b_M$ of width $2 \epsilon$. Then we can write the MoN Loss as
    \begin{align}
        L_N(P_T, P) \approx \sum_{i=1}^M P_T(b_i) \int_{b_i} \text{EMoN}_{P,b_i}(x^*) \diff x^* \label{eq:mon_loss_approx}
    \end{align}
    with 
    \begin{align}
        \text{EMoN}_{P,b_i}(x^*) = \int_{b_i} \min\left(|x^*-x_1|,|x^*-x_2|,...,|x^*-x_N|\right) \notag\\ P(x_1)P(x_2)\ldots P(x_N) \diff x_1 \diff x_2 \ldots \diff x_N
    \end{align}
    In expectation there are $N P(b_i)$ samples in bin $b_i$. Assume, that $z_i = N P(b_i)$ is an integer. Denote the sample that fall in $b_i$ as $H_i \coloneqq \{x_1^i, x_2^i,...,x_{n}^i\}$. Then, for the calculation of $ \text{EMoN}_{P,b_i}(x^*)$ we can ignore all samples that are not in $H_i$. Then we can write
    \begin{align}
        &\int_{b_i} \text{EMoN}_{P,b_i}(x^*) \diff x^* = \\& \int_{b_i} \int_{b_i} \min\left(|x^*-x_1^i|,|x^*-x_2^i|,...,|x^*-x_{z_i}^i|\right) \notag\\ & P(x_1^i)P(x_2^i)...P(x_{z_i}^i) \diff x_1^i \diff x_2^i,..., \diff x_{z_i}^i  \diff x^* \label{eq:double_integral}
    \end{align}
    
    For small $\epsilon$ and because we only consider differentiable functions, we can approximate $P(x_1^i)P(x_2^i)...P(x_{z_i}^i)$ withing the bin as uniform. Next we have to calculate the inner integral in \eqref{eq:double_integral}. For that,  center the bins around the origin (which is possible because we have uniform probability distributions) and consider the survival function of individual random samples $|x_i - x^*|$ \cite{overflow}:
   \begin{align}
        S(x)&=\Pr\{|x_i - x^*| > x\}  \\
        &= \Pr\{x_i > x^* + x\} + \Pr\{x_i < x^* - x\} \\
        &= \begin{cases}
            1 & \text {if } x \leq 0 \\
            1 - \frac{x}{\epsilon} & \text{if }  0 < x \leq \epsilon - |x^*|\\
            \frac{\epsilon + |x^*| - x}{2\epsilon} & \text {if } \epsilon - |x^*| < x < \epsilon + |x^*| \\
            0 & \text{if } x \geq \epsilon + |x^*|
        \end{cases}
    \end{align}

    Then the survival function $S_{MoN, z_i}(x)$ of $\min |x_i - x^*|$ is the probability, that all $z_i$ sample will independently be bigger than $x$. Thus the survival function is $S_{MoN, z_i}(x) = S(x)^{z_i}$ and therefore
    \begin{align}
        & E[\min|x_i - x^*|] = \\
        &= \int_0^{+\infty}S(x)^{z_i}dx \\
        &= \int_0^{\epsilon - |x^*|} \left(1 - \frac {x} {\epsilon} \right)^{z_i} dx
        + \int_{\epsilon - |x^*|}^{\epsilon + |x^*|} \left(\frac {\epsilon + |x^*| - x} {2\epsilon} \right)^{z_i} dx  \\
        &= \left. \frac {-\epsilon} {{z_i}+1}\left(1 - \frac {x} {\epsilon} \right)^{{z_i}+1} \right|_0^{\epsilon - |x^*|} + \left.\frac {-2\epsilon} {{z_i}+1} \left(\frac {\epsilon + |x^*| - x} {2\epsilon} \right)^{{z_i}+1} \right|_{\epsilon - |x^*|}^{\epsilon + |x^*|} \\
        &= \frac {\epsilon} {{z_i}+1}\left(1 - \frac {|x^*|^{{z_i}+1}} {\epsilon^{{z_i}+1}} + 2\frac {|x^*|^{{z_i}+1}} {\epsilon^{{z_i}+1}}\right) \\
        &= \frac {\epsilon} {{z_i}+1}\left(1 + \frac {|x^*|^{{z_i}+1}} {\epsilon^{{z_i}+1}} \right) \label{eq:expectation_min}
    \end{align}
     Substituting \eqref{eq:expectation_min} in \eqref{eq:mon_loss_approx} therefore yields
    \begin{align}
        L_N(P_T, P) & \approx \sum_{i=1}^M P_T(b_i) \int_{-\epsilon}^\epsilon \frac{\epsilon \left(1 + \frac {|x^*|^{z_i+1}} {\epsilon^{z_i+1}} \right)}{z_i+1} \diff x^* \\&=  
        \epsilon \sum_{i=1}^M P_T(b_i) \int_{-\epsilon}^\epsilon \frac{\left(1 + \frac {|x^*|^{z_i+1}} {\epsilon^{z_i+1}} \right)}{z_i+1} \diff x^* \\ &=
        \epsilon \sum_{i=1}^M P_T(b_i) \frac{2\epsilon + 2 \int_0^\epsilon \frac {{x^*}^{z_i+1}} {\epsilon^{z_i+1}} \diff x^*}{z_i+1}  \\ &=
        \underbrace{2 \epsilon^2}_{\coloneqq a} \sum_{i=1}^M P_T(b_i) \frac{1 + \frac{1}{NP(b_i) + 2} }{NP(b_i)+1}  \\ &\approx
        a \sum_{i=1}^M P_T(b_i) \frac{1 + \frac{1}{NP(b_i)} }{NP(b_i)} \label{eq:approx1}  \\ &=
        a \sum_{i=1}^M P_T(b_i)\left( \frac{1}{NP(b_i)} + \frac{1}{(NP(b_i))^2} \right)   \\ &\approx
        a \sum_{i=1}^M P_T(b_i) \frac{1}{NP(b_i)} \label{eq:approx2}
    \end{align} 
    where we used in line \eqref{eq:approx1} and \eqref{eq:approx2} that $N\rightarrow \infty$. We can find the minimum of $L_N(P_T, P)$ under the constraint that $\sum_{i=1}^M P(b_i) = 1$ by using a Lagrange multiplier. The objective is therefore:
    \begin{align}
        f(P(b_i), \lambda) & \coloneqq a \sum_{i=1}^M \frac{P_T(b_i)}{NP(b_i)} - \lambda \left(1 - \sum_{i=1}^M P(b_i)\right) 
    \end{align}
    Optimization for $P(b_i)$ yields:
    \begin{align}
        \nabla_{P(b_i)} f(P(b_i), \lambda) &= \lambda - a  \frac{P_T(b_i)}{N P(b_i)^2} \mbeq 0 \\
        \Leftrightarrow P(b_i) &= \sqrt{\frac{a P_T(b_i)}{\lambda N}} \label{condition1}\\
        \nabla_{\lambda} f(P(b_i), \lambda) &= 1 - \sum_{i=1}^M P(b_i) \mbeq 0 \\
        \Leftrightarrow \sum_{i=1}^M P(b_i) &= 1 \label{condition2}
    \end{align}
    where we can omit the $\pm$ in \eqref{condition1}, because probabilites can not be negative. Putting $\eqref{condition1}$ and  $\eqref{condition2}$ together results in:
    \begin{align}
        &\Rightarrow \sum_{i=1}^M P_T(b_i) \left( \sqrt{\frac{a P_T(b_i)}{\lambda N}} \right) = 1\\
        &\Rightarrow \lambda = \left( \frac{\sum_{i=1}^M \sqrt{a P_T(b_i)}}{\sqrt{N}}\right)^2 \\
        &\Rightarrow P(b_i) = \frac{\sqrt{P_T(b_i)}}{ \sum_{i=1}^M  \sqrt{P_T(b_i)}} \label{result}
    \end{align}
    Note, that this result is true for any distance metric in any dimensions, as long as the expected minimum distance of $N$ sample to a target sample goes locally with $\mathcal{O}(\frac{1}{N})$.
\end{proof}

\newpage

\begin{figure*}
    \centering
    \includegraphics[height=0.9\textheight]{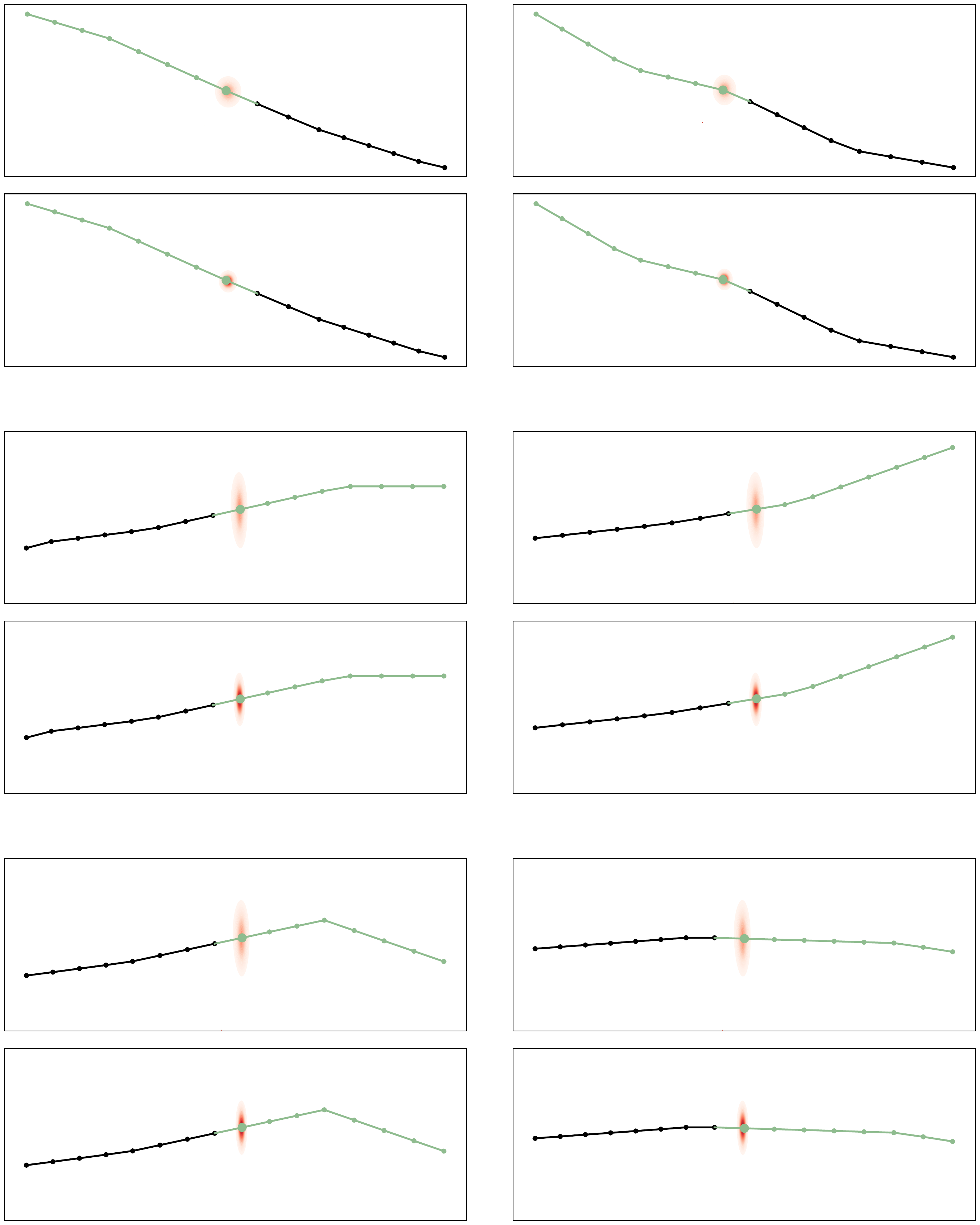}
    \caption{Marginalized probabilities of Social--GAN on the Zara dataset for $t=1$. Upper panel is the uncompensated PDF and the lower panel is the compensated one. Black is the observed trajectory and green the ground truth future.}
\end{figure*} 
\newpage
\clearpage
\FloatBarrier
\begin{figure*}
        \includegraphics[height=0.9\textheight]{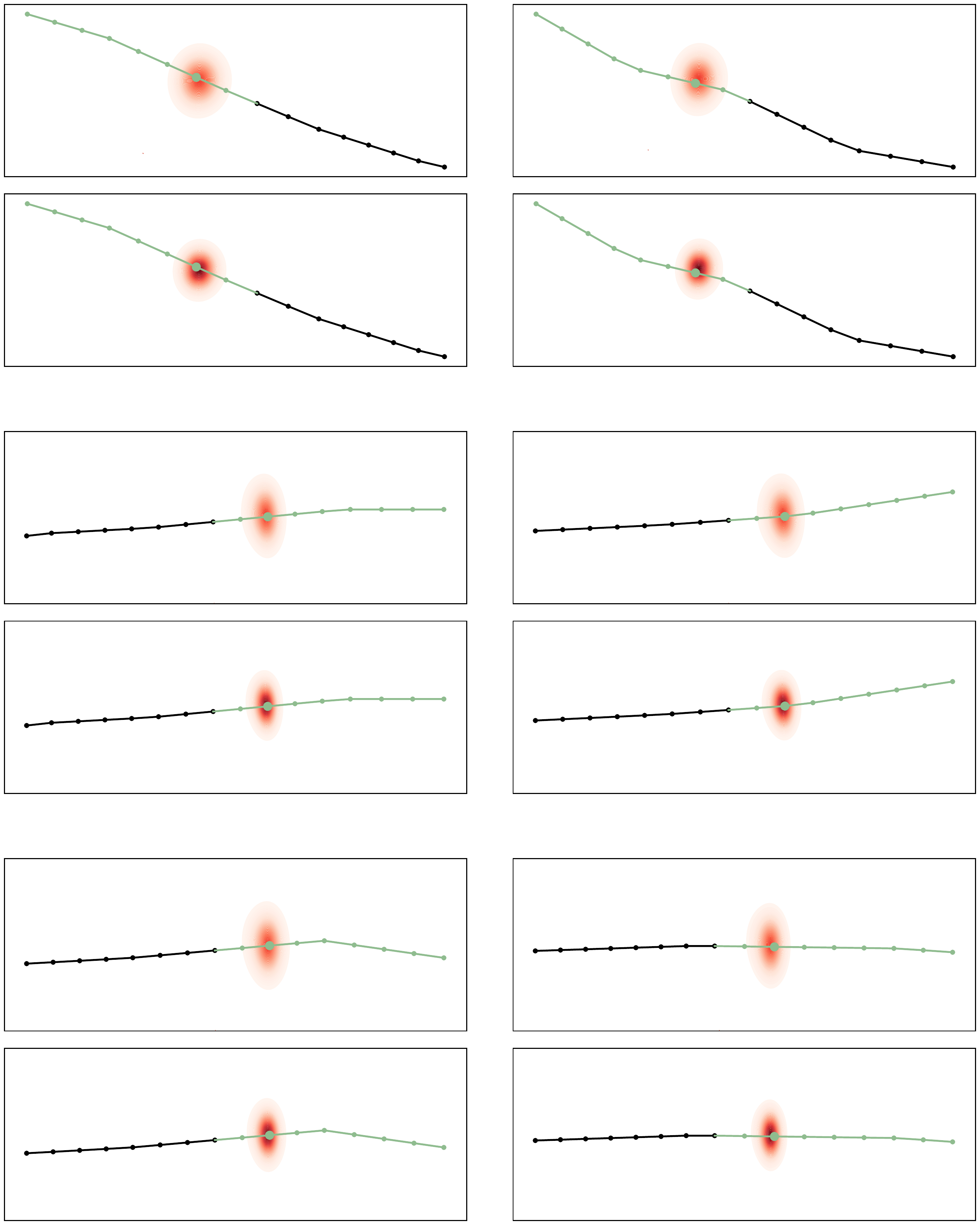}
        \caption{Marginalized probabilities of Social--GAN on the Zara dataset for $t=2$. Upper panel is the uncompensated PDF and the lower panel is the compensated one. Black is the observed trajectory and green the ground truth future.}
\end{figure*}
\newpage
\FloatBarrier
\begin{figure*}
        \includegraphics[height=0.9\textheight]{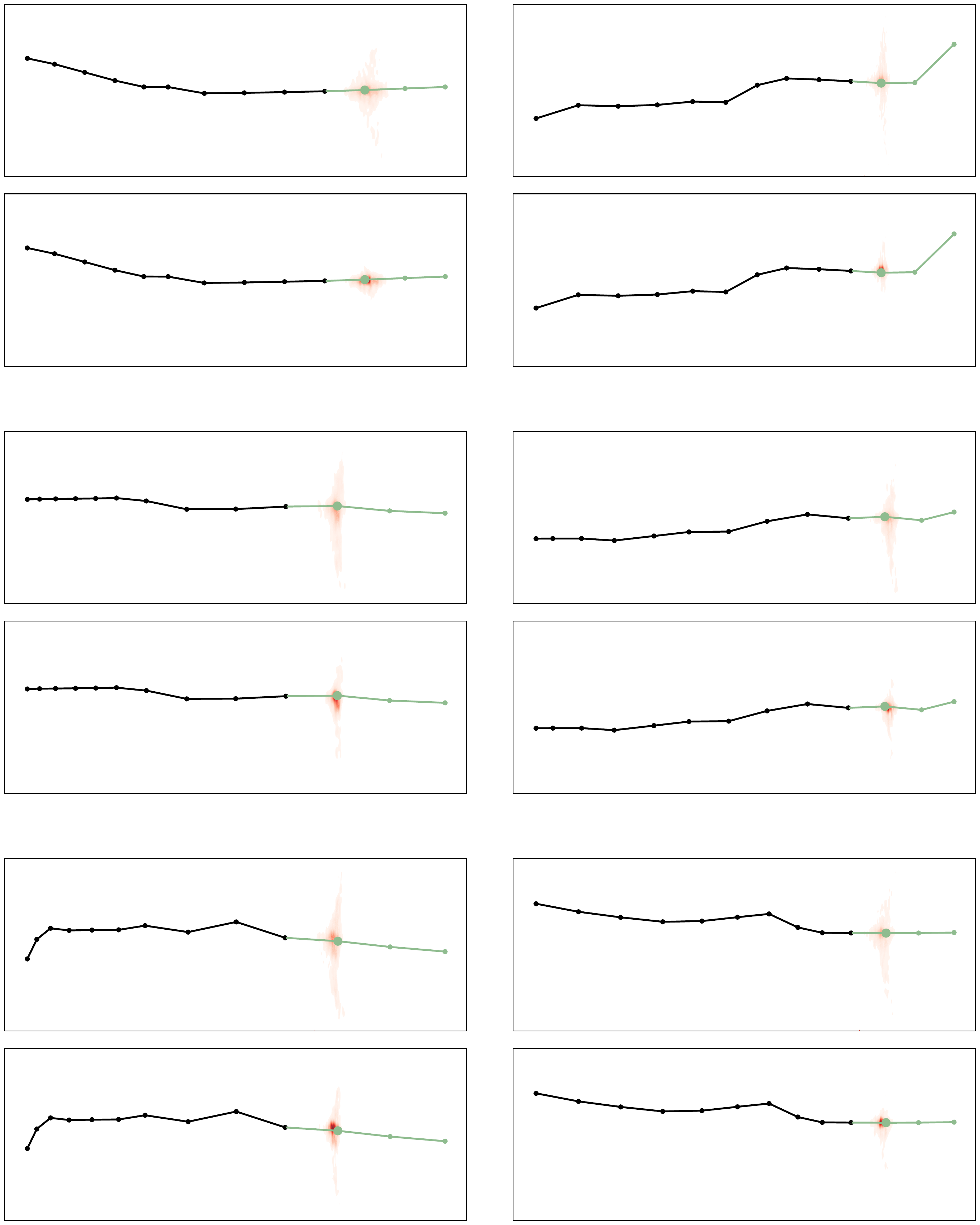}
        \caption{Marginalized probabilities of our own model on the NGSIM dataset for $t=1$. Upper panel is the uncompensated PDF and the lower panel is the compensated one. Black is the observed trajectory and green the ground truth future.}
\end{figure*} 
\newpage
\clearpage
\FloatBarrier
\begin{figure*}
        \includegraphics[height=0.9\textheight]{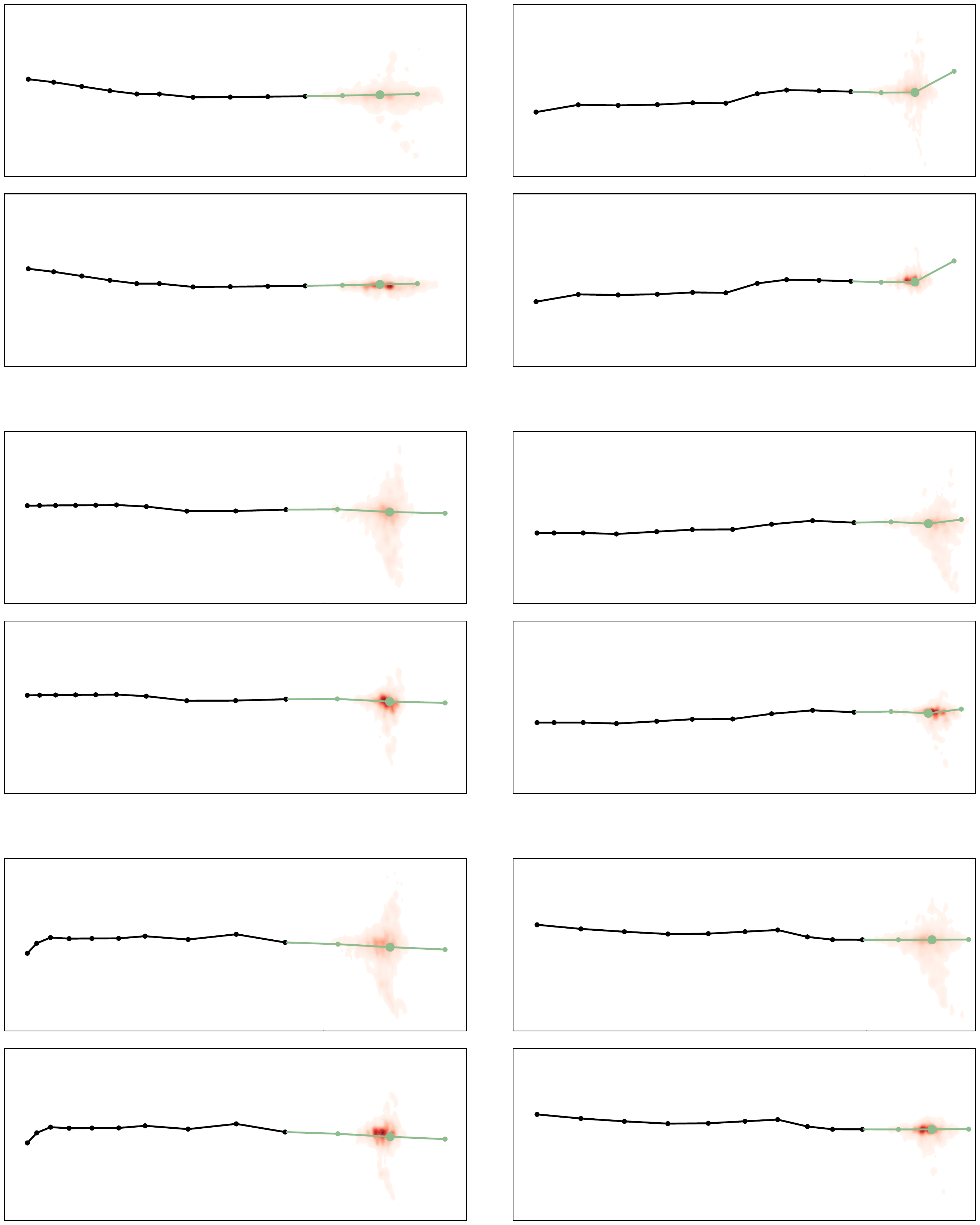}
        \caption{Marginalized probabilities of our own model on the NGSIM dataset for $t=2$. Upper panel is the uncompensated PDF and the lower panel is the compensated one. Black is the observed trajectory and green the ground truth future.}
\end{figure*}
\newpage
\clearpage
\FloatBarrier
\begin{figure*}
        \includegraphics[height=0.9\textheight]{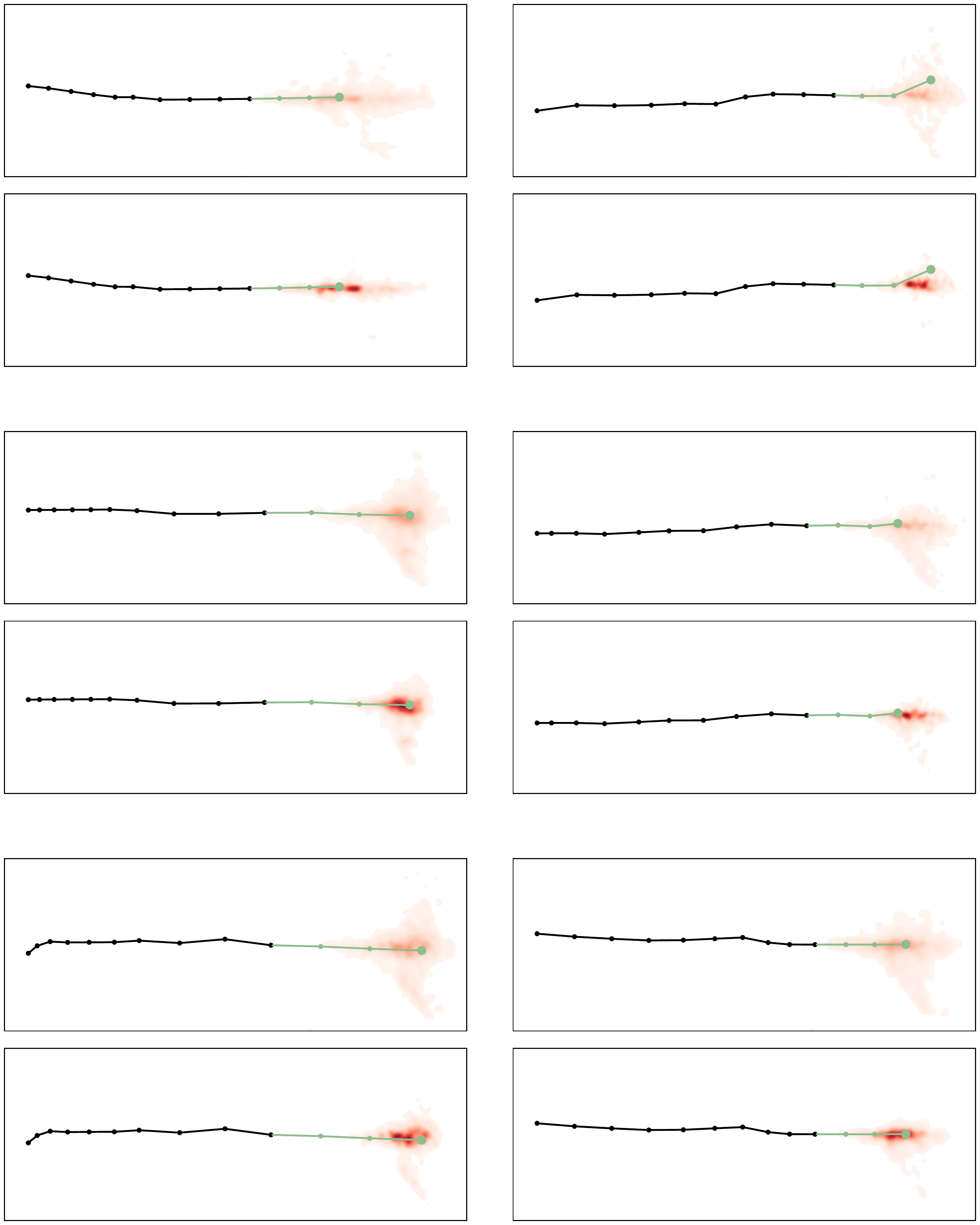}
        \caption{Marginalized probabilities of our own model on the NGSIM dataset for $t=3$. Upper panel is the uncompensated PDF and the lower panel is the compensated one. Black is the observed trajectory and green the ground truth future.}
\end{figure*}

\end{document}